\documentclass[10pt,twocolumn,letterpaper]{article}

\usepackage[pagenumbers]{cvpr} 
\usepackage{times}
\usepackage{caption}
\usepackage{epsfig}
\usepackage{graphicx}
\usepackage{amsmath}
\usepackage{amssymb}
\usepackage{bm}
\usepackage{cite}
\usepackage{booktabs}
\usepackage[accsupp]{axessibility}

\usepackage[dvipsnames]{xcolor}

\newcommand{\revised}[1]{#1}

\definecolor{cvprblue}{rgb}{0.21,0.49,0.74}
\usepackage[pagebackref,breaklinks,colorlinks,citecolor=cvprblue]{hyperref}

\def\paperID{11430} 
\def\confName{CVPR}
\def\confYear{2024}

\title{Makeup Prior Models for 3D Facial Makeup Estimation and Applications}

\author{Xingchao Yang$^{1,2}$, Takafumi Taketomi$^{1}$, Yuki Endo$^{2}$, Yoshihiro Kanamori$^{2}$\\
$^1$CyberAgent \quad$^2$University of Tsukuba\\
{\tt\small \{you\_koutyo, taketomi\_takafumi\}@cyberagent.co.jp, \{endo, kanamori\}@cs.tsukuba.ac.jp}
}

\begin{document}

\twocolumn[{%
\maketitle
\begin{center}
    \centering
    \captionsetup{type=figure}
    \includegraphics[width=\textwidth]{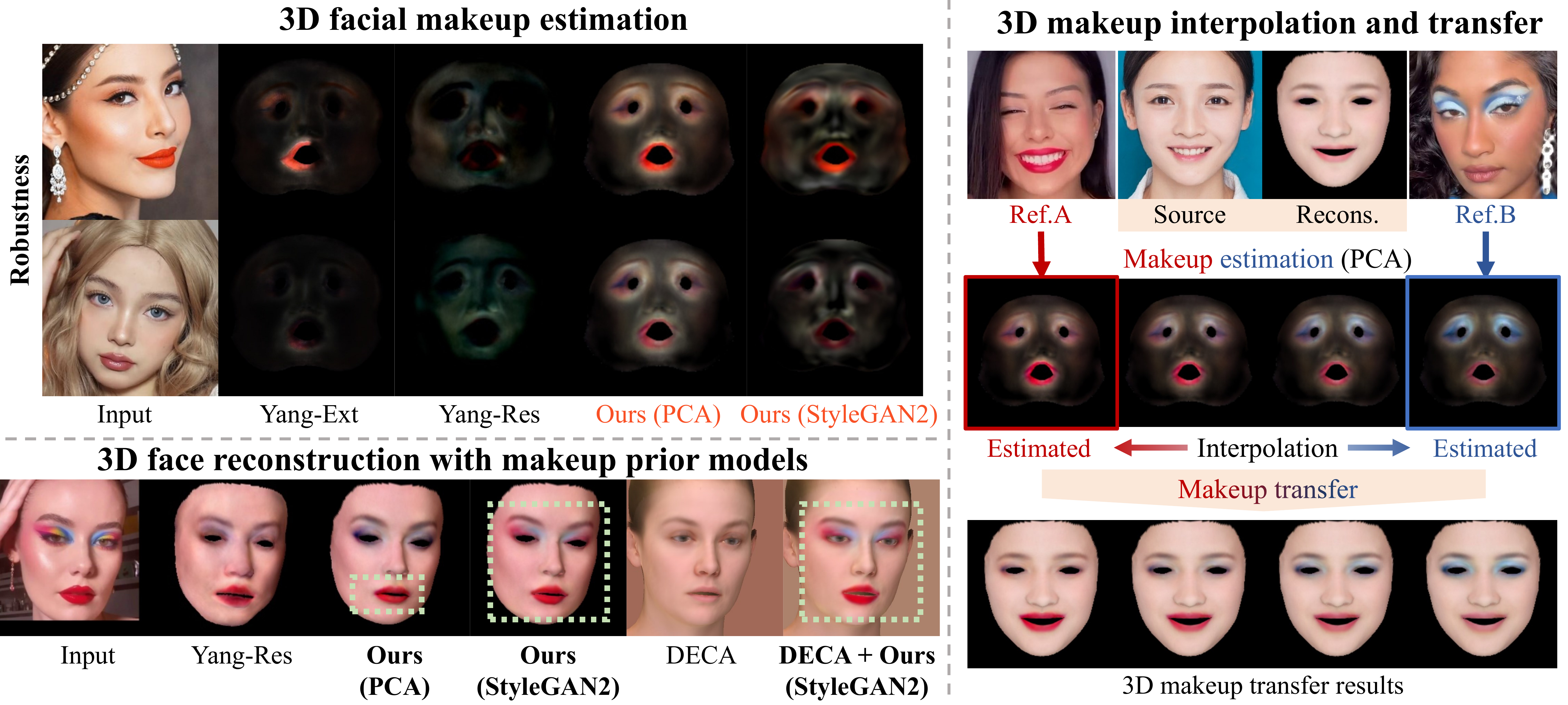}
    \captionof{figure}{\textbf{Example of 3D facial makeup estimation and applications using makeup prior models.} Top left: The effectiveness of our prior models (PCA and StyleGAN2) for estimating 3D facial makeup layers. Compared to the existing method~\cite{MakeupExtract} (Yang-Ext and Yang-Res), our method \revised{robustly} estimates makeup layers, \revised{especially in the case of self-occluded faces.} Bottom left: The result of 3D face reconstruction using makeup prior models. Our method accurately recovers the makeup of 3D faces and it can be compatible with the existing 3D face reconstruction framework~\cite{DECA}. Right: 3D makeup interpolation and transfer applications using the PCA-based prior model. Note that the StyleGAN2-based prior model has equivalent functionality.}
    \label{fig:teaser}
\end{center}%
}]

\begin{abstract}
In this work, we introduce two types of makeup prior models to extend existing 3D face prior models: PCA-based and StyleGAN2-based priors. The PCA-based prior model is a linear model that is easy to construct and is computationally efficient. However, it retains only low-frequency information. Conversely, the StyleGAN2-based model can represent high-frequency information with relatively higher computational cost than the PCA-based model. Although there is a trade-off between the two models, both are applicable to 3D facial makeup estimation and related applications. By leveraging makeup prior models and designing a makeup consistency module, we effectively address the challenges that previous methods faced in robustly estimating makeup, particularly in the context of handling self-occluded faces. In experiments, we demonstrate that our approach reduces computational costs by several orders of magnitude, achieving speeds up to 180 times faster. In addition, by improving the accuracy of the estimated makeup, we confirm that our methods are highly advantageous for various 3D facial makeup applications such as 3D makeup face reconstruction, user-friendly makeup editing, makeup transfer, and interpolation.
\end{abstract}

\begin{table*}[t]
\caption{\textbf{Comparative analysis of 3D makeup estimation and applications.} This table provides a functionalities comparison between our methods (PCA and StyleGAN2) and the existing techniques~\cite{MakeupExtract}. We evaluate various metrics including detail retention, \revised{robustness against self-occlusion}, processing speed for 3D makeup estimation, and capability in applications such as makeup transfer, interpolation, and user-friendly editing. Our methods demonstrate improved performance and versatility, particularly in challenging scenarios and application potential.}
\begin{center}
\resizebox{\textwidth}{!}{
{\small
\begin{tabular}{lccccccccc} 
\toprule
 & \multicolumn{3}{c}{\textbf{Makeup Layers}} & \multicolumn{3}{c}{\textbf{3D Makeup Estimation}} & \multicolumn{3}{c}{\textbf{Applications}} \\
\cmidrule(lr){2-4} \cmidrule(lr){5-7} \cmidrule(lr){8-10}
Functions & Types & Channels & Compositing & Details & \revised{Robustness} & Speed & Transfer & Interpolation & User-friendly editing \\
\midrule
Yang-Ext~\cite{MakeupExtract} & Textures & 4 & Alpha blend & $\checkmark$ & & 63.13s & $\checkmark$ & Alpha matte &\\
Yang-Res~\cite{MakeupExtract} & Prior & 3 & Residual & & & 56.12s & & &\\
\textbf{Ours (PCA)} & Prior & 4 & Alpha blend & & $\checkmark$ & 0.31s & $\checkmark$ & Alpha matte / coefficient & $\checkmark$\\
\textbf{Ours (StyleGAN2)} & Prior & 4 & Alpha blend & $\checkmark$ & $\checkmark$ & 18.13s & $\checkmark$ & Alpha matte / coefficient & $\checkmark$ \\
\bottomrule
\end{tabular}
}}
\end{center}
\label{tab:1}
\end{table*}

\section{Introduction}
\label{sec:intro}

The first 3D face prior model, known as the 3D Morphable Model (3DMM), was published over two decades ago~\cite{Blanz3DMM99}. In recent years, a diverse array of novel 3DMMs have been developed~\cite{bfm, LYHM, lsfm1, lsfm2, FLAME, CoMA, DiverseRaw, FML, AlbedoMM, FaceScape, CompleteMM, 3DMMfromSingle, FaceVerse, nonlinear3DMM, LearningPhyFace, i3DMM, headnerf, MoFaNeRF, NeuralHeadModel, ASM} and continue to play a crucial role in a wide range of face-related tasks~\cite{egger20203d, MonocularRTA, deng2020accurate, DECA, TRUST, DisCo, BEAT, clipface, DiffusionRig, DiFaReli}. To further extend and improve the functionality of 3DMMs, several methods are being developed that incorporate additional components such as self-shadowing illumination model and specular albedo~\cite{FaceWarehouse, IlluminationPrior, GlobalIllumiMM, AlbedoMM}. 
To further enhance the realism of the generated 3D faces, the latest methods employ Generative Adversarial Networks (GANs)~\cite{image2image, StyleGAN2} within the texture UV maps to construct a non-linear facial appearance prior model~\cite{saito2017photorealistic, HFFRI, GANFIT, LearningPhyFace, AvatarMe, AvatarMe++, UnsupervisedHFFG22, NormalizeAvatar, FitMe}.
However, the extension of 3DMMs to encompass makeup, an important element of human appearance, has not been widely discussed.

Makeup is a key component in creating realistic 3D digital humans in industries such as film, games, and advertising. 
Previous methods for capturing 3D facial makeup relied on direct measurement techniques that required specialized equipment~\cite{PBCR, SimulatingMakeup}. 
To facilitate easier access to makeup data, Yang \etal~\cite{MakeupExtract} propose a makeup texture extraction method (hereafter referred to as “Yang-Ext”). In addition, they represent the extracted makeup texture as a residual difference from the bare skin texture to construct a PCA-based linear prior model (hereafter referred to as “Yang-Res”). However, as shown in Tab.~\ref{tab:1}, their methods fall short in computational time, makeup estimation accuracy, and the diversity of applications. \revised{More details are explained in the
\textit{supplementary material}.}

In this paper, we \revised{build upon} the extracted makeup UV maps from Yang-Ext~\cite{MakeupExtract} to construct more sophisticated makeup prior models. 
One is a statistical linear model that can be efficiently constructed through Principal Component Analysis (PCA) and accompanied by a ResNet~\cite{ResNet} architecture for coefficient inference. This statistical approach is capable of capturing the global structure and color patterns of makeup, yet it may fall short in reproducing the nuanced details, such as the gradient of eyeshadows. 
The other is a generative nonlinear prior model using StyleGAN2~\cite{StyleGAN2}. The StyleGAN2-based prior model requires more computational resources and always involves a carefully designed GAN inversion method to search for the desired coefficient. We incorporate a pSp Encoder~\cite{pSp} and follow it with a minor fine-tuning phase, inspired by~\cite{NormalizeAvatar, endoCAVW2022}. Although resource-intensive, this StyleGAN2-based approach excels at realistically reproducing intricate makeup details in images.
\revised{The primary advantages of our makeup prior models lie in their reduced computational time and improved robustness in makeup estimation, as summarized in Tab.~\ref{tab:1}. The improvements are attributed to our use of a regression-based approach, as well as the design of a dedicated makeup consistency module and corresponding regularization loss functions.} Furthermore, it demonstrates the versatility of various 3D facial makeup applications, as illustrated in Fig.~\ref{fig:teaser}.

In summary, our contributions are:
\begin{itemize}
    \setlength{\itemsep}{0.2em}
    \setlength{\parskip}{0.2em}
    \item Developing two makeup prior models - a PCA-based linear model and a StyleGAN2-based generative model - each with a tailored coefficient estimation network;
    \item \revised{Integrating the makeup consistency module with corresponding regularization loss functions, our specialized network architecture not only boosts robustness in makeup estimation but also minimizes computational time;}
    \item Demonstrating applicability in advanced 3D makeup-related applications such as 3D makeup face reconstruction, user-friendly editing, makeup transfer, and interpolation.
\end{itemize}


\section{Related Work}

\subsection{Facial Makeup Estimation}
3D facial makeup significantly elevates the realism and charm of virtual digital humans. Nevertheless, obtaining 3D facial makeup remains a formidable task~\cite{PBCR, CGF11Makeup}. Traditionally, this process demands considerable time and effort from artists, especially in designing shaders and materials for makeup. Therefore, research that aims to rapidly and accurately acquire 3D facial makeup from the real world proves beneficial.

Inspired by 2D makeup transfer methods~\cite{beautyGAN, PSGAN, ladn, CPM}, Yang \etal~\cite{MakeupExtract} introduced a pioneering method for 3D facial makeup acquisition: step-by-step texture decomposition-based extraction (Yang-Ext). Additionally, they utilize the extracted makeup textures to build a PCA-based makeup prior model and then apply an analysis-by-synthesis approach to estimate the makeup residual relative to bare skin (Yang-Res). However, the practical application of these makeup estimation methods is still challenging. It requires significant computational time and \revised{presents a challenge for self-occluded faces.} 
Moreover, due to Yang-Res's approach of over-fitting residual makeup, it is applicable primarily to 3D face reconstruction and not suitable for other applications like makeup transfer~\cite{MakeStar, SimulatingMakeup, PairedCycleGAN, beautyGAN, BeautyGlow, PSGAN, ladn, CPM, SCGAN, SOGAN, elegant, RamGAN, BeautyREC}.

We address these limitations by developing effective makeup prior models and introducing corresponding makeup estimation network architectures. Our approach significantly improves the accuracy and reductions in inference time, while also expanding the potential for a wide range of applications.

\subsection{Morphable Face Prior Model}
The 3D Morphable Model (3DMM) is a statistical prior model that utilizes a dataset of known 3D shapes and textures~\cite{Blanz3DMM99}. Leveraging 3DMM enables a multitude of facial-related applications, such as face recognition, reconstruction, tracking, and animation~\cite{egger20203d, MonocularRTA}. To capture the extensive diversity in facial shapes and textures across different ethnicities and ages, numerous enhancements have been made to the original 3DMM concept~\cite{bfm, LYHM, lsfm1, lsfm2, FLAME, CoMA, DiverseRaw, AlbedoMM, FaceScape, CompleteMM, FaceVerse, LearningPhyFace, ASM}, including developments such as the Basel Face Model (BFM)~\cite{bfm} and FLAME~\cite{FLAME}. 

To enhance the utility and applicability of 3DMMs, researchers have introduced various methods such as facial expression modeling~\cite{FaceWarehouse} and advanced lighting techniques~\cite{IlluminationPrior, GlobalIllumiMM}. The AlbedoMM~\cite{AlbedoMM} proposes a specular albedo prior model to better account for specular reflections. Yang \etal~\cite{MakeupExtract} introduces a residual makeup prior model (Yang-Res). Since traditional statistical linear models mainly capture broad patterns and lack fine details, modern approaches~\cite{saito2017photorealistic, HFFRI, uvgan, GANFIT, AvatarMe, LearningPhyFace, AvatarMe++, Bao:2021:ToG, FitMe, UnsupervisedHFFG22, THFFSOR, OSTeC, dsd-gan, NormalizeAvatar, bareskinnet, MakeupExtract, FFHQ-UV, SSIF, clipface} utilize deep generative models~\cite{image2image, StyleGAN2} to recover detailed 3D facial appearance materials. In this paper, we use 3D facial makeup textures extracted from 2D images by Yang-Ext~\cite{MakeupExtract} to build two types of makeup prior models, each with its own advantages and limitations. The first is a statistical PCA model, which excels in speed, while the second is an advanced StyleGAN2 generative model, noted for its high-quality output. Our models show improved performance and functionality compared to the Yang-Res model.

\subsection{Model-based 3D Face Reconstruction}
Model-based approaches for 3D face reconstruction focus on deriving optimal 3DMM coefficients and rendering the corresponding 3DMM to match an input image~\cite{reconstruction, PixelRec, Face2Face}. With the advent of deep learning, regression-based methods~\cite{RegressionDeep, mofa, CNNRecons, unsupervised3dmm, 250Hz, deng2020accurate, DECA, INORig, CEST, EMOCA, MICA, TRUST, FOCUS, AccurateTransformer} have gained prominence, offering more stable and quicker reconstruction outcomes, as detailed in various comprehensive surveys~\cite{MonocularRTA, egger20203d}. In this paper, our makeup estimation network is based on a 3D face reconstruction task and introduces a makeup consistency module to enhance the quality of estimated makeup. This module facilitates the separation of makeup from facial identity and expressions, making it well-suited for applications such as makeup interpolation and transfer.


\section{Approach}
We first describe our makeup prior models (Sec.~\ref{sec:models}) and then explain the makeup estimation network (Sec.~\ref{sec:makeup_estimate_net}). Finally, we introduce an improvement method for the StyleGAN2 makeup prior model (Sec.~\ref{sec:StyleGAN2_improv}).

\begin{figure}[t]
    \centering
    \includegraphics[width=0.8\linewidth]{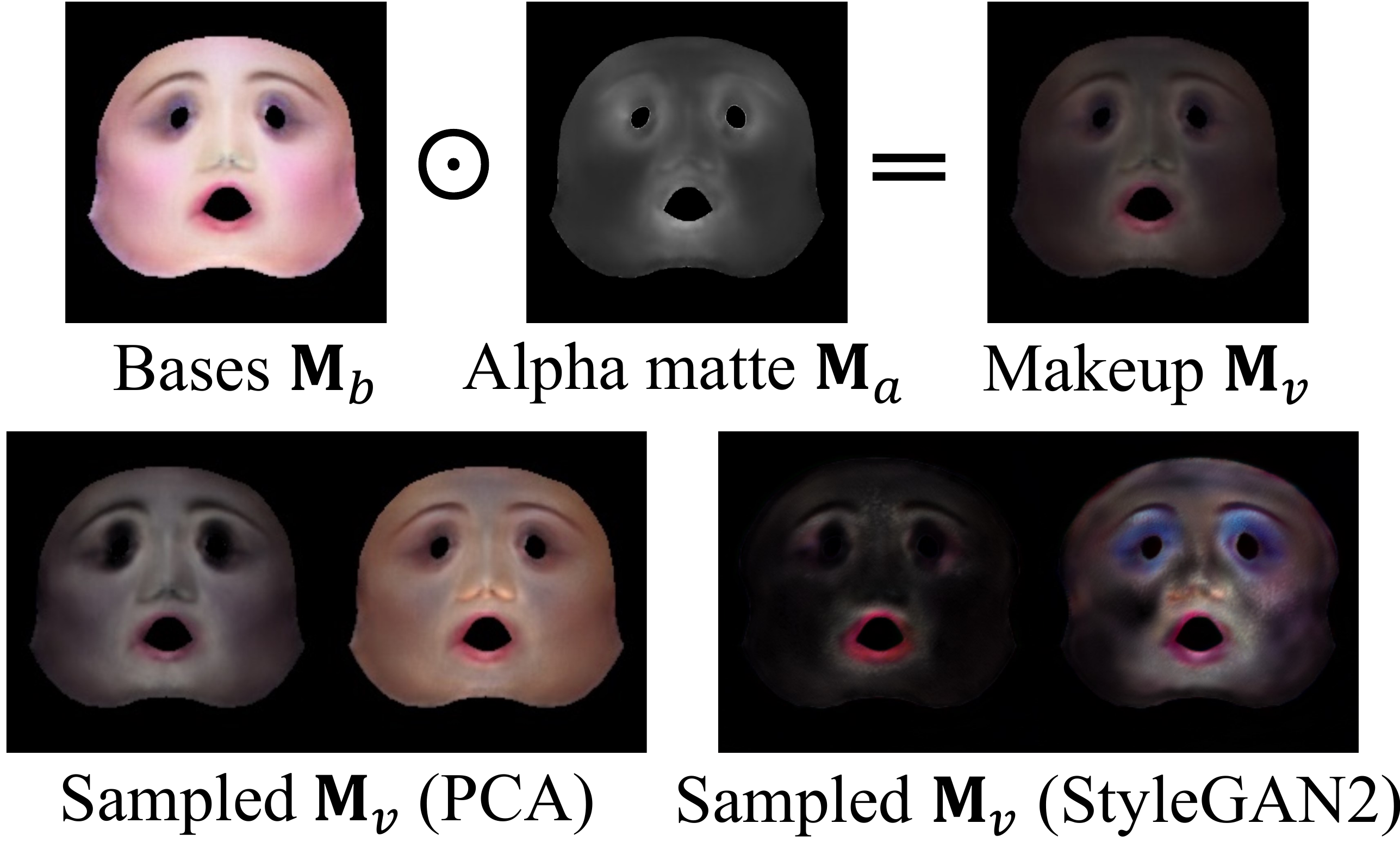}
    \caption{\textbf{Examples of makeup sampled from PCA and StyleGAN2 prior models.} The top row demonstrates visual composites $\mathbf{M}_{v}$ from the combination of makeup bases $\mathbf{M}_{b}$ and alpha matte $\mathbf{M}_{a}$. The bottom row shows variations produced by manipulating the coefficient in the prior models.
}
\label{fig:models}
\end{figure}

\subsection{Makeup Prior Models}
\label{sec:models}
Fig.~\ref{fig:models} presents examples of makeup and an alpha matte sampled from the prior models.
The makeup textures $\mathbf{M} \in\mathbb{R}^{D}$ ($D = d{\times}d{\times}4$ and $d$ is the UV map size) generated by the prior model (PCA or StyleGAN2)  consist of the 4-channel tuple (r, g, b, a) for each pixel in the UV map, where the first three channels are makeup bases $\mathbf{M}_b \in \mathbb{R}^{d \times d \times 3}$ and the last channel is the corresponding alpha matte $\mathbf{M}_{\alpha} \in \mathbb{R}^{d \times d}$. The pixel values of $\mathbf{M}_a$ range from 0 to 1.
We use the visually intuitive representation $\mathbf{M}_{v} \in \mathbb{R}^{d \times d \times 3}$ to display makeup, providing a clear depiction of its color and pattern, calculated as follows:
\begin{align}
&\mathbf{M}_{v} = \mathbf{M}_{b} \odot \mathbf{M}_{a},
\label{eq:makeup_vis}
\end{align}
where ${\odot}$ denotes the element-wise multiplication. 

\vskip0.5\baselineskip
\noindent \textbf{PCA-based Prior Model.}
We utilize PCA to construct a linear parametric makeup prior model $\mathbf{M}(\mathbf{\upsilon})\in\mathbb{R}^{D}$ as follows:
\begin{align}
&\mathbf{M} = \bar{\mathbf{M}} + \mathbf{B}_{m}\mathbf{\upsilon},
\label{eq:pca_M}
\end{align}
where $\bar{\mathbf{M}}\in\mathbb{R}^{D}$ is the average facial makeup, $\mathbf{B}_{m}\in\mathbb{R}^{D{\times}100}$ is the PCA bases, and $\mathbf{\upsilon}\in\mathbb{R}^{100}$ is the corresponding coefficient vector for adjusting 3D facial makeup. 

In contrast, Yang-Res~\cite{MakeupExtract} constructs its model using $\mathbf{M}_{v}$ as a makeup residual, resulting in a 3-channel statistical model. Consequently, their model lacks a dedicated channel for alpha matte information.

\vskip0.5\baselineskip
\noindent \textbf{StyleGAN2-based Prior Model.}
\label{sec:StyleGAN2-prior}
We train StyleGAN2~\cite{StyleGAN2}, with normally distributed noise as input, to construct a generative non-linear makeup prior model with a latent vector \( \mathbf{w} \in \mathcal{W}{+} \) in the latent space. The disentangled latent space for this StyleGAN2 model is defined as \( \mathcal{W}{+}:= \mathbb{R}^{7 \times 512} \).

\begin{figure}[t]
    \centering
    \includegraphics[width=\linewidth]{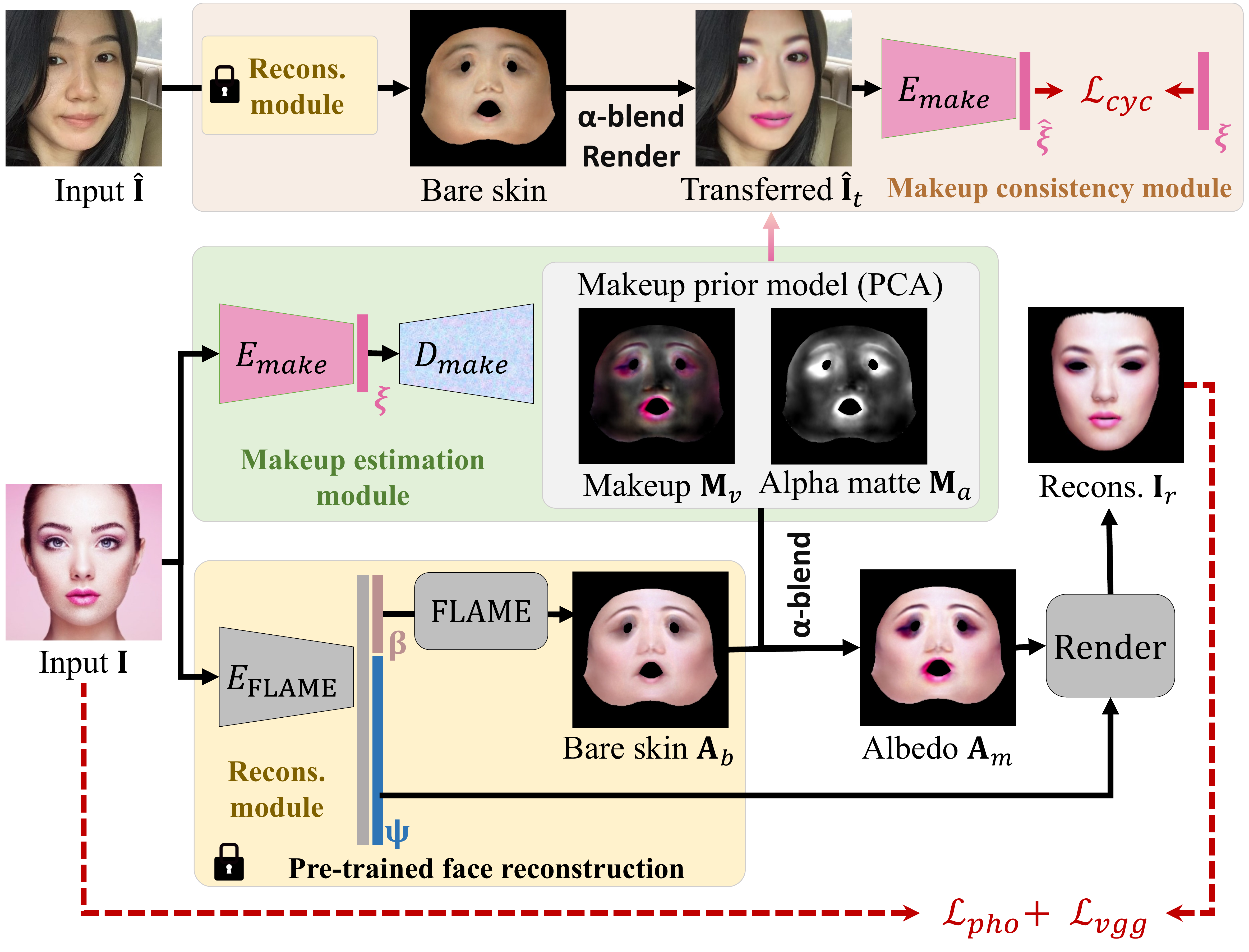}
    \caption{\textbf{Overview of the makeup estimation network architecture.} The network is composed of three modules: The Reconstruction module is pre-trained for 3D face reconstruction; The makeup estimation module employs ${E}_{make}$ to infer the makeup coefficient $\mathbf{\xi}$ and generates associated makeup textures; The makeup consistency module enhances the effectiveness of makeup estimation.}
    \label{fig:method}
\end{figure}

\subsection{Makeup Estimation Network}
\label{sec:makeup_estimate_net}

Our network is designed primarily for makeup estimation within the context of 3D face reconstruction. It incorporates a newly developed makeup consistency module and is trained in a self-supervised learning manner. As depicted in Fig.~\ref{fig:method}, the network consists of three modules: the Reconstruction module, the makeup estimation module, and the makeup consistency module. The combination of the Reconstruction module and the makeup estimation module forms the core of the 3D face reconstruction functionality, while the makeup consistency module enhances the effectiveness of the makeup estimation process.

\noindent \textbf{Reconstruction module.} For the basic 3DMM model that does not include makeup, we use the FLAME model~\cite{FLAME}, resulting in the estimated albedo $\mathbf{A}_{b}$ referred to as “bare skin”. We adopt \revised{the same} pre-trained encoder ${E}_\text{FLAME}$ and \revised{rendering methods} from the existing method~\cite{MakeupExtract} to infer FLAME coefficients $\beta$ for bare skin and $\psi$  for rendering (\ie, geometry, camera parameters, and illumination conditions). We freeze ${E}_\text{FLAME}$ during training our entire network.

\noindent \textbf{Makeup estimation module.} In the makeup estimation module, the choice of an encoder $E_{make}$ and a decoder $D_{make}$ depends on the employed prior model, that is, PCA or StyleGAN2. The encoder ${E}_{make}$ has a role for inferring coefficients $\xi$ ($\upsilon$ or $\mathbf{w}$) of our makeup prior model. (1) Specifically, for the \textbf{PCA}-based makeup prior model, the encoder ${E}_{make}$ employs the ResNet50~\cite{ResNet} architecture to infer the coefficients $\mathbf{\upsilon}$. The decoder ${D}_{make}$ follows Eq.~(\ref{eq:pca_M}) to generate makeup textures. (2) For the \textbf{StyleGAN2}-based makeup prior model, we use the pSp Encoder~\cite{pSp} as ${E}_{make}$ to infer the coefficients $\mathbf{w}$. The decoder uses the pre-trained StyleGAN2 makeup prior model (Sec.~\ref{sec:StyleGAN2-prior}) as ${D}_{make}$ and removes the noise injection during generation to achieve precise control.

The final albedo $\mathbf{A}_{m}$ containing makeup is calculated via alpha blending as follows:
\begin{equation}
\begin{aligned}
\mathbf{A}_{m} &= \mathbf{M}_{b} \odot \mathbf{M}_{a}  + (1 - \mathbf{M}_{a}) \odot \mathbf{A}_{b}\\
&= \mathbf{M}_{v} + (1 - \mathbf{M}_{a}) \odot \mathbf{A}_{b},
\label{eq:blending}
\end{aligned}
\end{equation}
where $(1 - \mathbf{M}_{a})$ refers to the inversion of $\mathbf{M}_{a}$.
We use differentiable rendering to reconstruct 3D face $\mathbf{I}_{r}$ from $\mathbf{A}_{m}$ and $\psi$. Subsequently, the encoder $E_{make}$ is trained by minimizing the error between the original image $\mathbf{I}$ and $\mathbf{I}_{r}$ to ensure the makeup similarity.

\begin{figure}[t]
    \centering
    \includegraphics[width=0.99\linewidth]{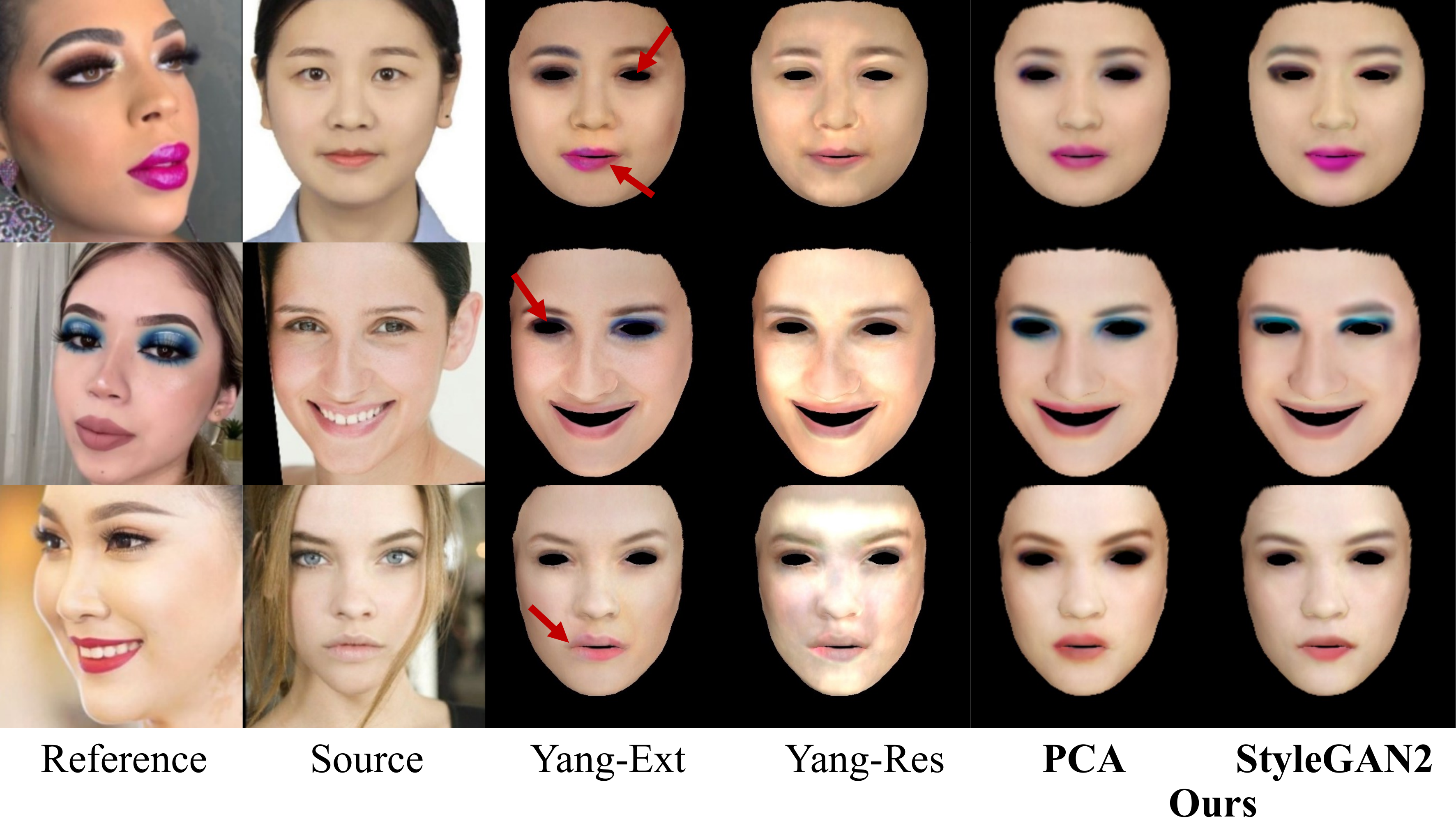}
    \caption{\textbf{Comparison of 3D makeup transfer.} The results confirm that Yang-Ext~\cite{MakeupExtract} struggles with large pose variations and Yang-Res~\cite{MakeupExtract} is constrained by its dependency on reference images, our method demonstrates accurate and reliable makeup transfer.}
    \label{fig:result_transfer}
\end{figure}

Note that the characteristics of the estimated makeup layers as summarized in Tab.~\ref{tab:1}, the difference between $\mathbf{A}_{m}$ in Eq.~(\ref{eq:blending}) and the makeup-applied albedo in Yang-Res~\cite{MakeupExtract} (denoted as $\mathbf{A}'_m$). In the Yang-Res method, the makeup layer is represented as the residual (denoted as $\mathbf{M}_{\Delta}$) between the bare skin and the input image. As shown in Fig.~\ref{fig:result_transfer}, it becomes challenging when applied to makeup transfer. $\mathbf{A}'_m$ is calculated as follows:
\begin{align}
\mathbf{A}'_{m} = \mathbf{A}_{b}  + \mathbf{M}_{\Delta}.
\label{eq:addition}
\end{align}

\noindent \textbf{Makeup consistency module.}
This module aims to mitigate over-fitting in makeup estimation from input image $\mathbf{I}$, accounting for variations in pose, lighting, and expression. Estimated makeup obtained from the coefficient $\xi$ is applied to randomly selected bare faces $\mathbf{\hat{I}}$. The encoder ${E}_{make}$ then re-estimates makeup on the transferred image $\mathbf{\hat{I}}_t$ to estimate makeup coefficient $\hat{\xi}$, ensuring that the new coefficient $\hat{\xi}$ consistent with the original coefficient $\xi$. This module is not required during the inference phase.

\noindent \textbf{Loss function.}
The final loss function is defined as follows:
\begin{equation}
\begin{aligned}
\mathcal{L}(\mathbf{\xi}) &= \lambda_{pho} \, \mathcal{L}_{pho}(\mathbf{\xi}) + \lambda_{vgg} \, \mathcal{L}_{vgg}(\mathbf{\xi}) \\
                          &+ \lambda_{cyc} \, \mathcal{L}_{cyc}(\mathbf{\xi}) + \lambda_{reg} \, \mathcal{L}_{reg}(\mathbf{\xi}) ,
\end{aligned}
\end{equation}
where $\mathcal{L}_{pho}$ represents the L1 loss, and $\mathcal{L}_{vgg}$ represents the perceptual loss~\cite{vggloss}. $\mathcal{L}_{cyc}$ represents the MSE loss between makeup coefficients $\hat{\xi}$ and $\xi$. $\mathcal{L}_{reg}$ is a regularization term, which uses the L2 norm to prevent the inferred parameters from deviating too much from the mean value.
The weight coefficients are tuned to achieve a balance between the different losses and are set to
$\lambda_{pho}=100$, 
$\lambda_{vgg}=1$, 
$\lambda_{cyc}=4$ for PCA, 
$\lambda_{cyc}=20$ for StyleGAN2, and $\lambda_{reg}=1 \times 10^{-4}$.

\subsection{Refinement for StyleGAN2 Prior Model}
\label{sec:StyleGAN2_improv}
We observe that the StyleGAN2-based approach occasionally yields imprecise results when inferring less common makeup styles, which we attribute to limitations in the training data leading to out-of-domain problems~\cite{GANInversion}.
To overcome this issue, we follow methodologies from~\cite{NormalizeAvatar, endoCAVW2022} to perform minor optimization-based fine-tuning on the estimated coefficient $\mathbf{w}$. We remove the makeup consistency module and its associated loss function $\mathcal{L}_{cyc}$ from $\mathcal{L}_{\xi}$, while other loss functions are preserved.
Alternatively, we optimize the coefficient using two additional loss functions to constrain the generated makeup textures. The first is a soft symmetry loss $\mathcal{L}_{sym}$, ensuring bilateral symmetry in $\mathbf{M}$, and the second is a regulation loss $\mathcal{L}'_{reg}$ for $\mathbf{M}_{a}$. This regulation loss acts as a preventive measure against over-fitting that may occur when the proportion of makeup is excessive. The additional loss functions can be formalized as follows:
\begin{equation}
\begin{aligned}
\mathcal{L}'(\mathbf{\mathbf{w}}) &= \lambda_{pho} \, \mathcal{L}_{pho}(\mathbf{\mathbf{w}}) + \lambda_{vgg} \, \mathcal{L}_{vgg}(\mathbf{\mathbf{w}}) \\
                          &+ \lambda_{reg} \, \mathcal{L}_{reg}(\mathbf{\mathbf{w}}) + \lambda'_{reg} \, \mathcal{L}'_{reg}(\mathbf{\mathbf{w}})  \\
                          &+ \lambda_{sym} \, \mathcal{L}_{sym}(\mathbf{\mathbf{w}}),
\end{aligned}
\end{equation}
where $\mathcal{L}_{sym}$ is the L1 loss and $\mathcal{L}'_{reg}$ regularizes alpha matte using $\Vert{\mathbf{M}_{a}}\Vert_{1}$, which represents the L1 norm. The weight parameters are set to $\lambda_{sym} = 8$, and $\lambda'_{reg} = 1$.

\begin{figure}[t]
    \centering
    \includegraphics[width=0.99\linewidth]{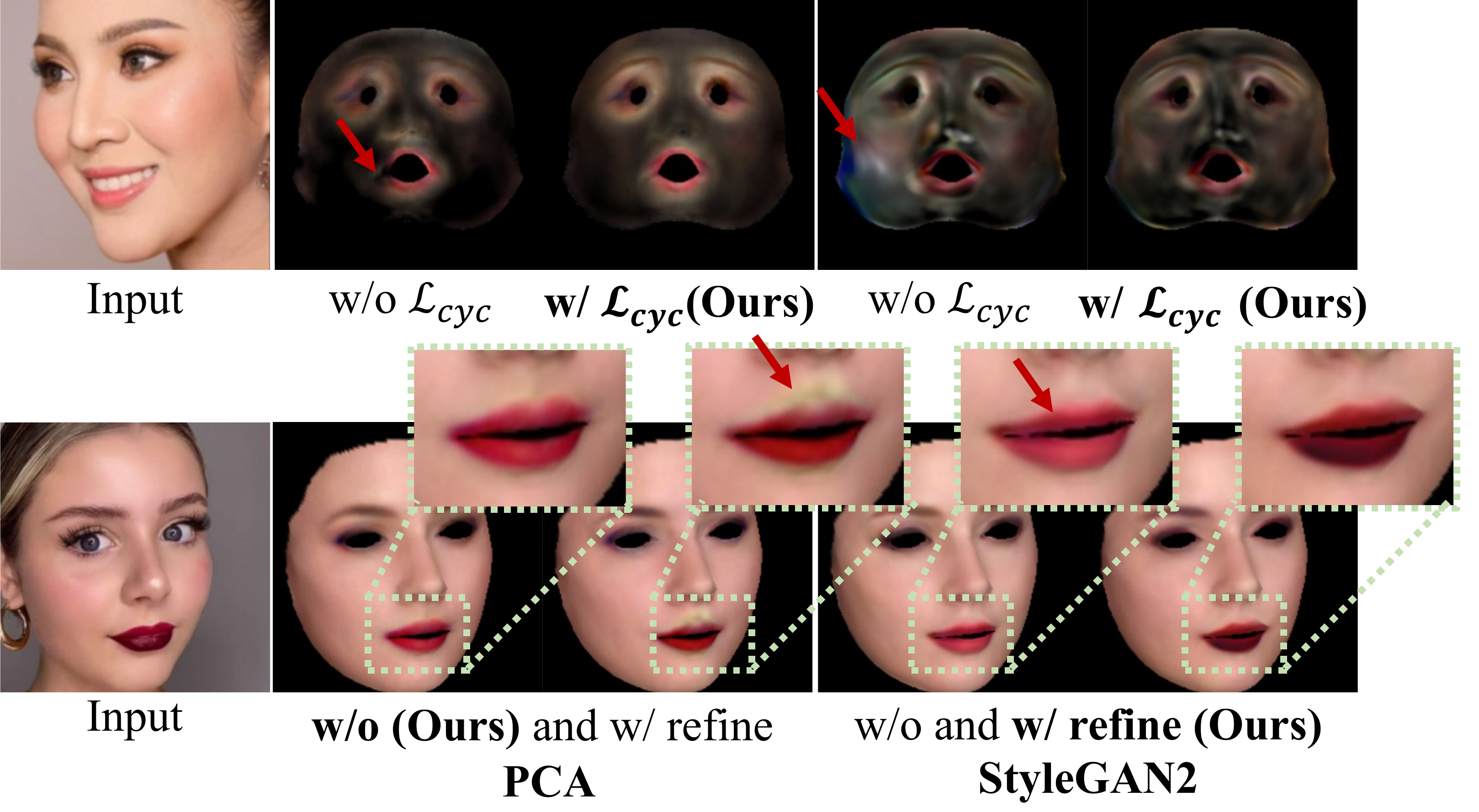}
    \caption{\textbf{Qualitative comparison for ablation study.} The top row demonstrates that incorporating the makeup consistency module contributes to the robustness of the results. The bottom row reveals that while refinement introduces artifacts in the PCA model, it significantly improves the accuracy of the StyleGAN2 model, as evidenced by the more precise makeup application on the lips.}
    \label{fig:ablation}
\end{figure}

\begin{figure}[t]
    \centering
    \includegraphics[width=0.99\linewidth]{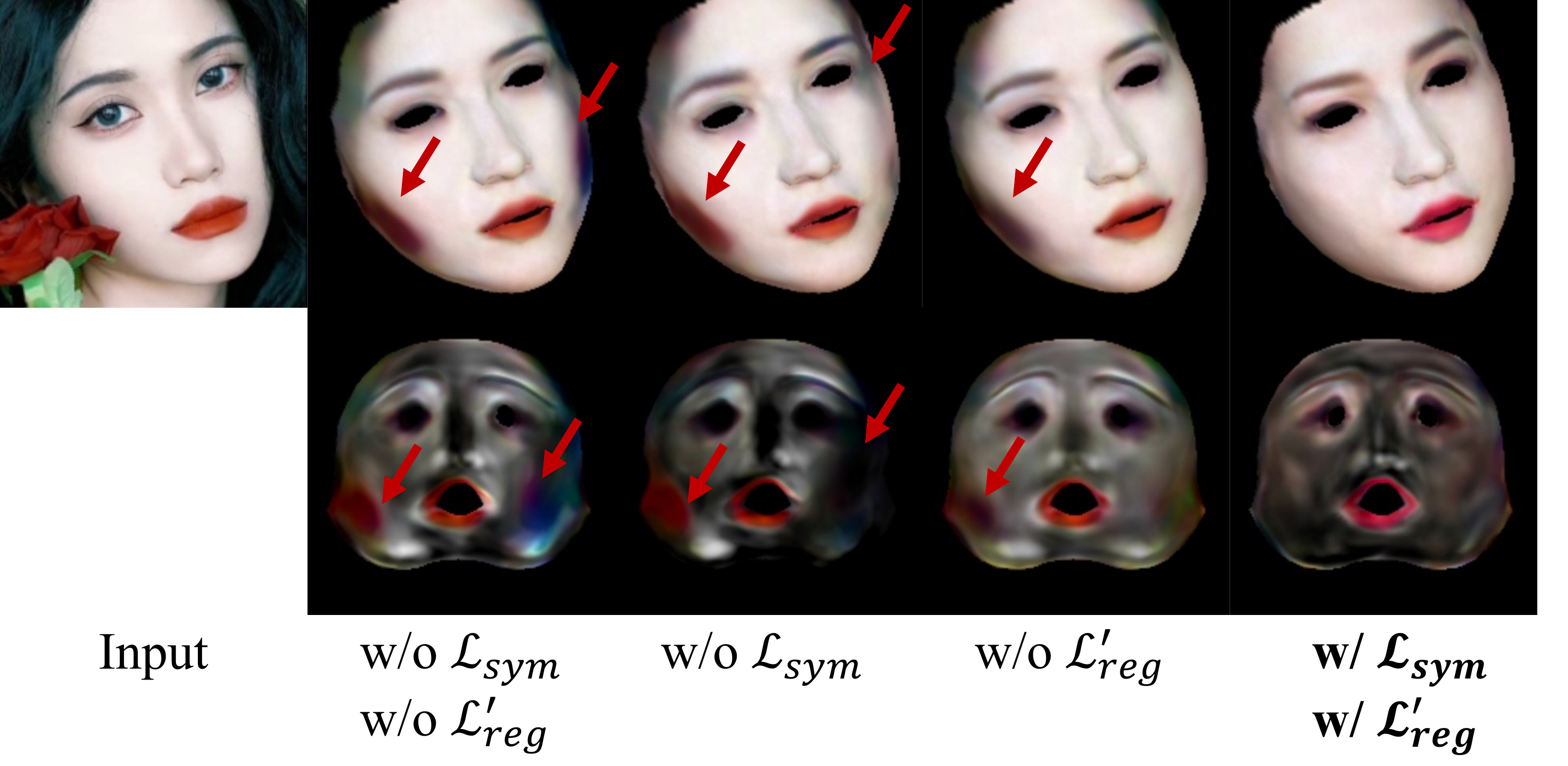}
    \caption{\textbf{Effect of loss functions.} In the refinement process for our StyleGAN2 model, without loss functions $\mathcal{L}_{sym}$ and $\mathcal{L}'_{reg}$ leads to over-fitting of the input image, resulting in artifacts.}
    \label{fig:ablation_refine}
\end{figure}

In our experiments with the PCA-based makeup prior model, we also attempted refinement through fine-tuning. Unfortunately, we found that minor adjustments failed to yield significant improvements, while extensive perturbation of the PCA coefficients led to the generation of numerous artifacts. We hypothesize the limitations of the PCA model are reached with complex makeup, as its lower degree of freedom in the model offers limited scope for optimization. Therefore, we conduct the refinement process to the StyleGAN2 model, which demonstrates enhanced results. The ablation comparison results confirm the necessity and effectiveness as shown in Figs.~\ref{fig:ablation} and \ref{fig:ablation_refine}.


\begin{figure*}[t]
    \centering
    \includegraphics[width=0.95\linewidth]{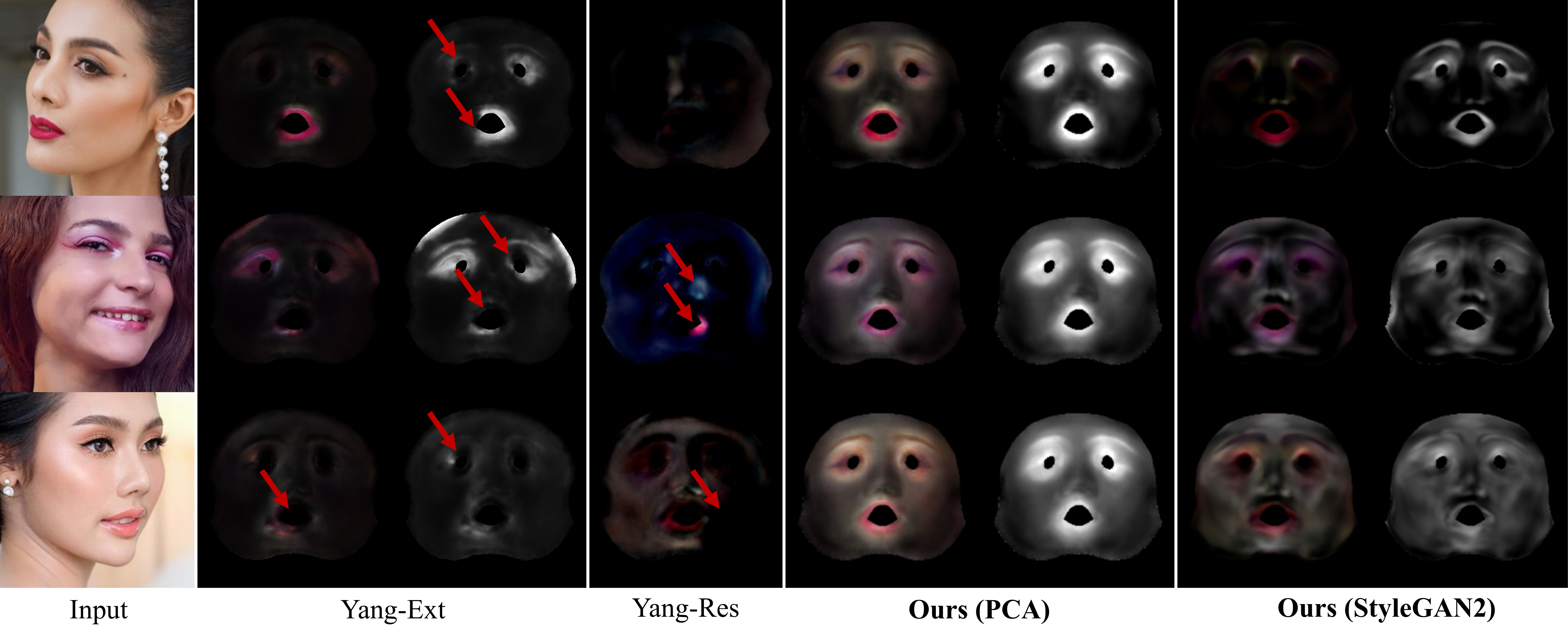}
    \caption{\textbf{Comparison with 3D facial makeup estimation methods.} The results demonstrate the robustness of our methods (PCA and StyleGAN2) in terms of stability and accuracy in estimating makeup, outperforming both Yang-Ext and Yang-Res~\cite{MakeupExtract}, which show limitations \revised{in handling self-occluded faces.}}
    \label{fig:result_estimation}
\end{figure*}

\label{sec:quantitative}
\begin{table*}
\caption{\textbf{Comparative analysis of 3DMM-based 3D face reconstruction on Wild~\cite{PSGAN} and BeautyFace~\cite{BeautyREC} datasets.} Values in \textbf{bold} represent the best results, while those \underline{underlined} denote the second-best. $^*$ denotes the integration of makeup prior models.}
\begin{center}
\resizebox{0.95\textwidth}{!}{
{\small
\begin{tabular}{lcccccccccc}
\toprule
\multicolumn{1}{l}{ } & \multicolumn{5}{c}{\textbf{Wild}} & \multicolumn{5}{c}{\textbf{BeautyFace}} \\
\cmidrule(lr){2-6} \cmidrule(lr){7-11}
Method & HM(eyes)$\downarrow$ & HM(lips)$\downarrow$ & RMSE$\downarrow$ & SSIM$\uparrow$ & LPIPS$\downarrow$ & HM(eyes)$\downarrow$ & HM(lips)$\downarrow$ & RMSE$\downarrow$ & SSIM$\uparrow$ & LPIPS$\downarrow$\\ 
\midrule
DECA~\cite{DECA} & 0.0048 & 0.0088 & 0.0876 & 0.3651 & 0.0871 & 0.0043 & 0.0106 & 0.0636 & 0.4002 & 0.0840\\
DECA~\cite{DECA}\textbf{* (PCA)} & \underline{0.0045} & \underline{0.0081} & \underline{0.0871} & \textbf{0.3683} & \underline{0.0822} & \underline{0.0041} & \underline{0.0081} & \underline{0.0595} & \underline{0.4041} & \underline{0.0779}\\
DECA~\cite{DECA}\textbf{* (StyleGAN2)} & \textbf{0.0044} & \textbf{0.0076} & \textbf{0.0807} & \underline{0.3678} & \textbf{0.0798} & \textbf{0.0039} & \textbf{0.0077} & \textbf{0.0568} & \textbf{0.4042} & \textbf{0.0747}\\
\midrule
Reconstruction module & 0.0045 & 0.0093 & 0.0667 & 0.5940 & 0.0750 & 0.0037 & 0.0102 & 0.0666 & 0.5034 & 0.0745\\
Yang-Res~\cite{MakeupExtract} & \underline{0.0037} & 0.0079 & \underline{0.0543} & 0.6036 & \underline{0.0674} & \underline{0.0032} & 0.0081 & \textbf{0.0545} & \underline{0.5065} & \underline{0.0680}\\
\textbf{Ours (PCA)} & 0.0041 & \underline{0.0078} & 0.0609 & \underline{0.6111} & 0.0681 & 0.0035 & \underline{0.0078} & 0.0690 & 0.5013 & 0.0733\\
w/o Refine (StyleGAN2) & 0.0042 & 0.0083 & 0.0618 & 0.6091 & 0.0685 & 0.0035 & 0.0084 & 0.0695 & 0.5031 & 0.0741\\
\textbf{Ours (StyleGAN2)} & \textbf{0.0036} & \textbf{0.0073} & \textbf{0.0517} & \textbf{0.6240} & \textbf{0.0608} & \textbf{0.0031} & \textbf{0.0068} & \underline{0.0650} & \textbf{0.5134} & \textbf{0.0673}\\
\bottomrule
\end{tabular}
}}
\end{center}
\label{tab:2}
\end{table*}

\begin{figure}[t]
    \centering
    \includegraphics[width=0.99\linewidth]{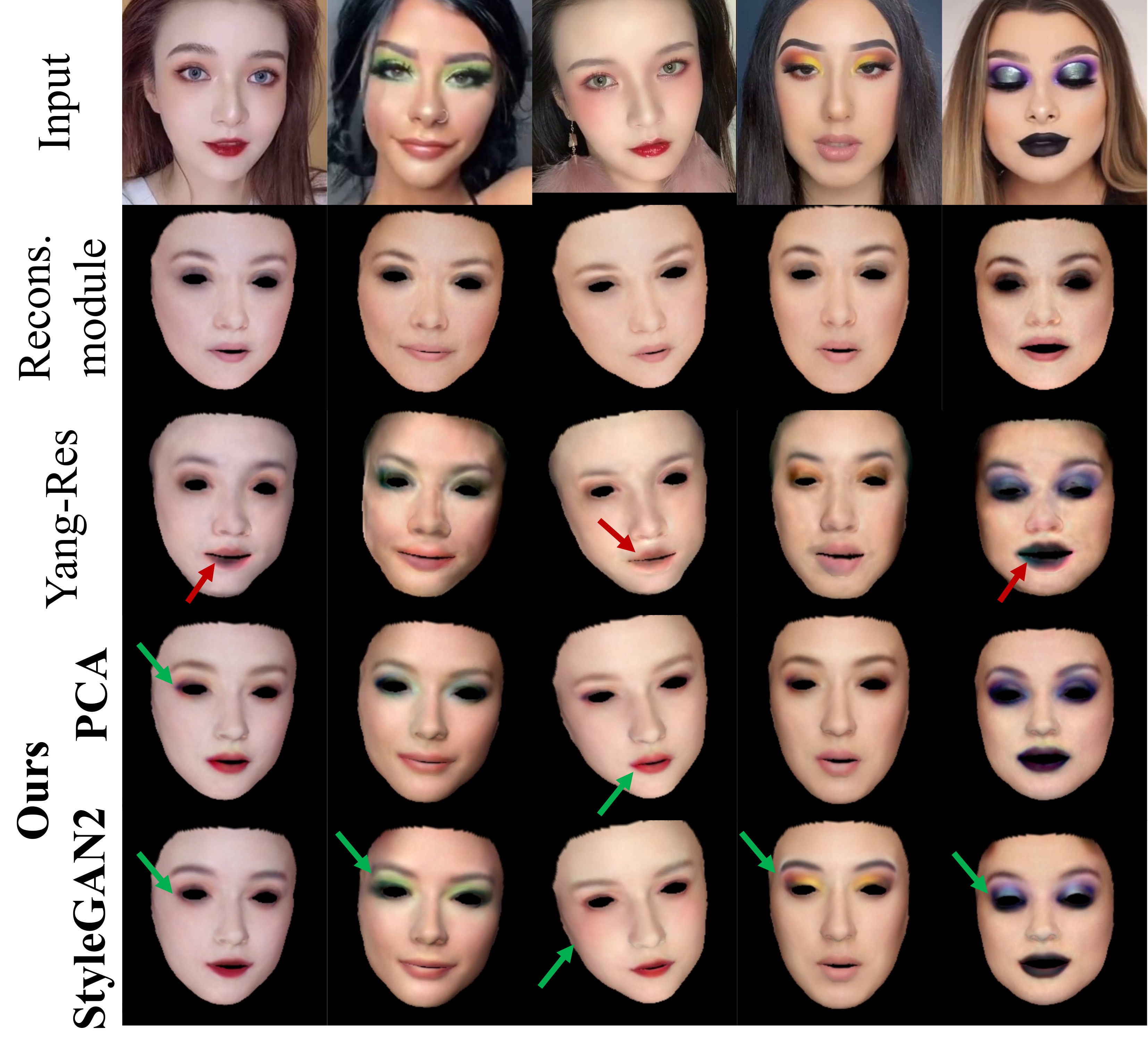}
    \caption{\textbf{Comparison with 3D face reconstruction methods using makeup prior models.} Our methods successfully reconstruct facial makeup. Specifically, the PCA model is capable of broadly recovering makeup colors, while the StyleGAN2 model achieves precise replication of complex makeup features, such as blush and gradient eyeshadow.}
    \label{fig:result_reconstruction}
\end{figure}

\begin{figure}[t]
    \centering
    \includegraphics[width=0.99\linewidth]{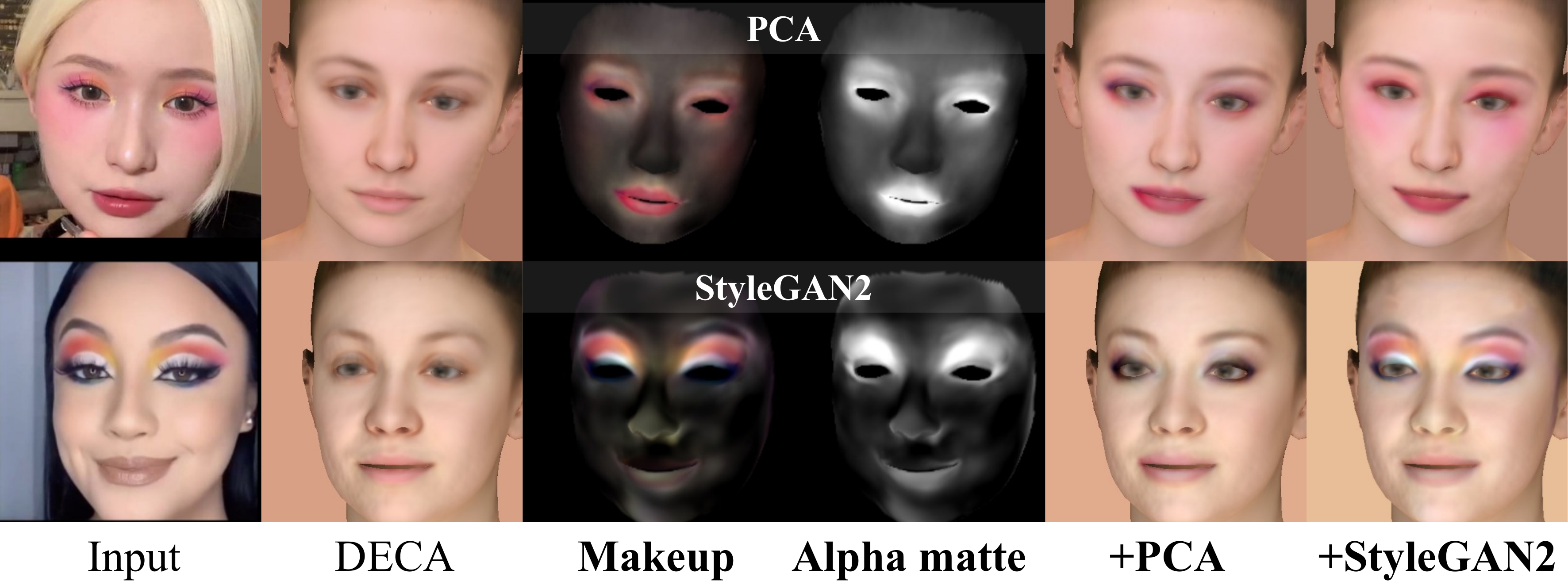}
    \caption{\textbf{Combining DECA and our makeup prior models.} Our makeup prior models can be incorporated into model-based 3D face reconstruction methods with the same topology, thereby endowing the capability to handle facial makeup.}
    \label{fig:result_deca_recons}
\end{figure}

\begin{figure}[t]
    \centering
    \includegraphics[width=0.99\linewidth]{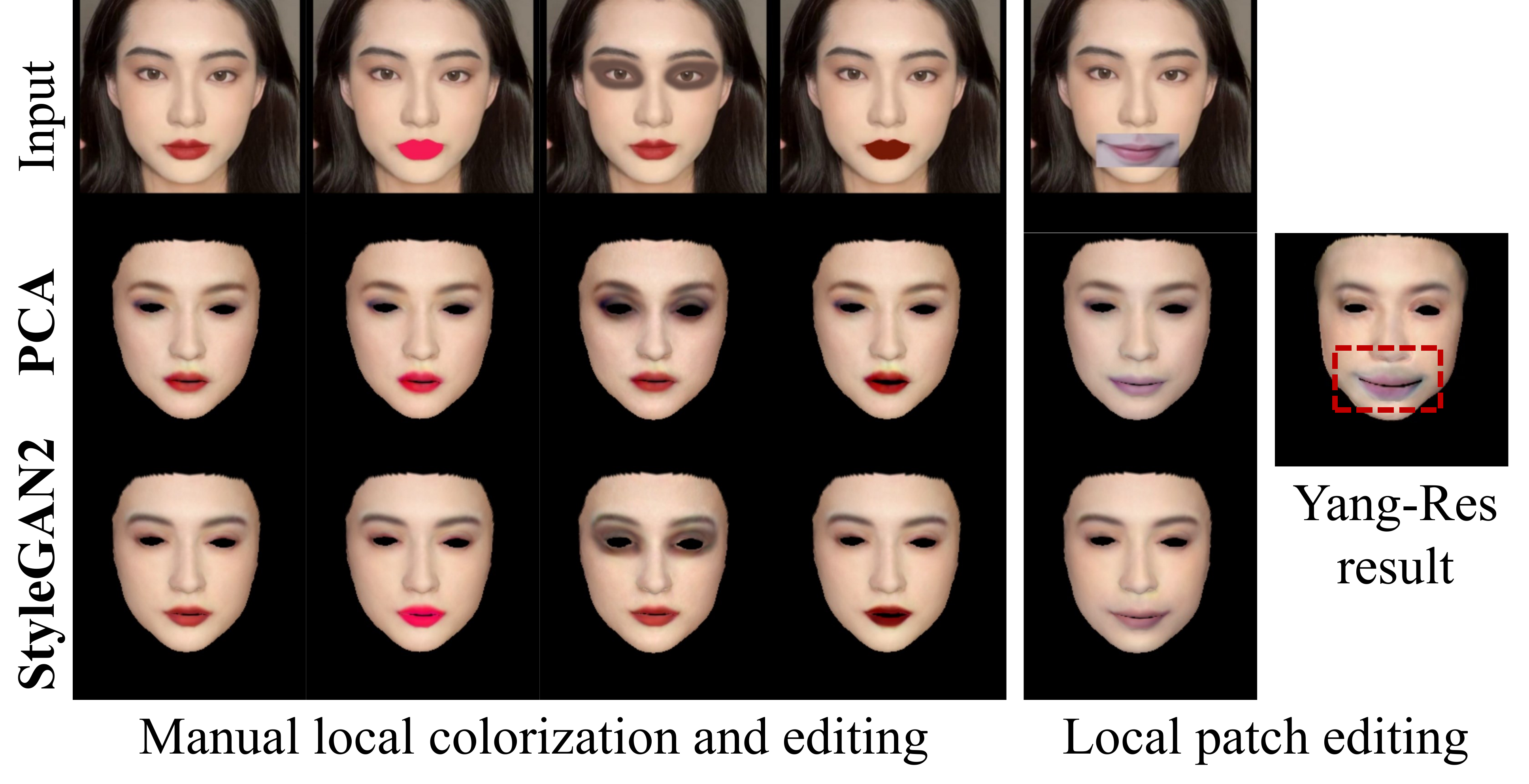}
    \caption{\textbf{Robustness and ease for makeup editing.} The results validate our method's robustness and ease in achieving stable 3D face reconstruction with makeup edits for input images, effectively handling color changes and local patch editing. The Yang-Res approach is impacted by over-fitting issues.}
    \label{fig:result_edit}
\end{figure}

\begin{figure}[t]
    \centering
    \includegraphics[width=0.99\linewidth]{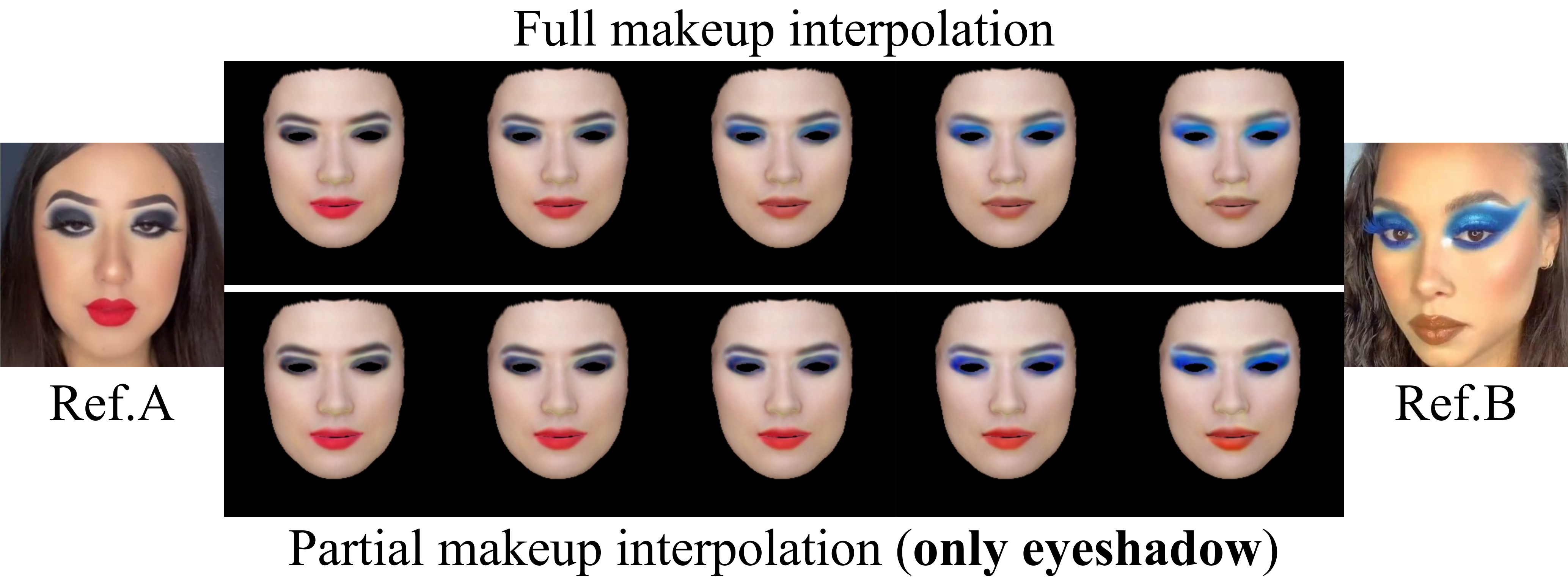}
    \caption{\textbf{Results of 3D makeup interpolation.} 3D makeup interpolation experiments utilizing the StyleGAN2 prior model. The top row demonstrates full interpolation between two makeup styles, blending both eyeshadow and lipstick. The bottom row highlights the capability of StyleGAN2 makeup prior model for partial makeup interpolation and transfer.}
    \label{fig:result_interpolate}
\end{figure}

\section{Implementation Details}
\noindent \textbf{Makeup prior models.}
Our prior models (PCA and StyleGAN2) are both constructed using 3,070 images of makeup bases and alpha matte, which are extracted by Yang-Ext~\cite{MakeupExtract} following the UV format of FLAME model~\cite{FLAME}. These materials are sourced from makeup datasets~\cite{beautyGAN, ladn}. The UV map size is set to $d = 256$, following the configuration of FLAME, and only the facial area is utilized.
The training of the StyleGAN2 makeup prior model requires a total duration of ten days using NVIDIA L4 GPU. To ensure stability during training, we retain the noise injection mechanism~\cite{noise_injection}, while it is omitted during the inference phase.

\vskip0.5\baselineskip
\noindent \textbf{Networks.}
The makeup estimation network is trained using the same dataset~\cite{beautyGAN, ladn} for building the makeup prior models. We set the batch size to 10 for the PCA model and 4 for the StyleGAN2 model. We use the Adam optimizer with the learning rate of $1 \times 10^{-4}$.  The model is trained for 20 epochs using NVIDIA 2080 Ti GPU. The network training requires one day for the PCA model and two days for the StyleGAN2 model. For the refinement of the StyleGAN2 model, we configure the learning rate to $1 \times 10^{-2}$ for 40 iterations.
The computation times for inference are discussed in Sec.~\ref{sec:computation_time}.

\section{Experiments}
The summary of functional comparisons is presented in Tab.~\ref{tab:1}. We utilize makeup datasets Wild~\cite{PSGAN} and BeautyFace~\cite{BeautyREC} for experiments, which are not employed in the construction of the makeup prior model or in the network training. These datasets comprise a substantial number of images featuring challenging facial poses, expressions, and complex makeup, amounting to 380 and 2,995 images, respectively. 
We evaluate the makeup prior models, through a diverse set of experiments, such as makeup estimation, 3D face reconstruction, user-friendly makeup editing, makeup transfer, interpolation, and computational time. 
In the experiments, the red arrows denote results with deficiencies, while green arrows indicate favorable outcomes. 

\subsection{3D Facial Makeup Estimation}
Before discussing the various 3D facial makeup applications, we first assess the quality of makeup estimation. As shown in Figs.~\ref{fig:teaser} and \ref{fig:result_estimation}, Yang-Ext~\cite{MakeupExtract} faces challenges in estimating makeup on \revised{self-occluded faces}. Furthermore, the step-by-step image decomposition of their method results in unstable makeup estimation, often missing details such as the color and region of the upper lip, an issue that is also present in the Yang-Res~\cite{MakeupExtract}. In contrast, our method consistently estimates the makeup regions and can handle makeup on facial areas not visible in the image. Compared to the results of the PCA prior model, the StyleGAN2 prior model captures a greater level of detail. Our effective makeup estimation facilitates subsequent 3D facial makeup applications.

\subsection{Applications}
\vskip0.5\baselineskip
\noindent \textbf{3D face reconstruction.}
Fig.~\ref{fig:result_reconstruction} displays comparative results of model-based 3D facial reconstruction with progressively increasing makeup complexity from left to right. Traditional 3DMM-based methods lack the capability to reconstruct faces with makeup, while the Yang-Res~\cite{MakeupExtract} often converges to local minima, resulting in 3D faces with artifacts. In contrast, our methods demonstrate effective reconstruction of facial makeup. The PCA prior model approximately restores the makeup colors, whereas the StyleGAN2 prior model accurately recovers challenging makeup features, such as blush and gradient eyeshadow.

Our makeup prior models are compatible with other 3DMM-based reconstruction methods (require the same topology), allowing for seamless integration without additional training. Figs.~\ref{fig:teaser} and \ref{fig:result_deca_recons} demonstrate the combined efficacy of our PCA and StyleGAN2 estimated makeup with the DECA~\cite{DECA} framework.

To measure the difference between the reconstructed faces and the original images, we conduct a quantitative evaluation using several metrics, including Histogram Matching (HM), Root Mean Square Error (RMSE), Structural Similarity Index Measure (SSIM), and Learned Perceptual Image Patch Similarity (LPIPS)~\cite{lpips} metrics. The HM metric measures the similarity of the color distributions between two images and is commonly used in makeup analysis and transfer. We perform segmentation around the eyes and lips to compare the HM metric regionally, thereby assessing the reconstruction quality of eyeshadow and lipstick. The metrics presented in Tab.~\ref{tab:2} indicate that both the PCA model and the StyleGAN2 model enhance reconstruction precision, with StyleGAN2 being superior. Detailed quantitative comparisons, including ablation results of loss functions, will be provided in the supplementary materials.

\vskip0.5\baselineskip
\noindent \textbf{User-friendly makeup editing.}
Fig.~\ref{fig:result_edit} demonstrates the reconstructed results after editing the makeup in the original images. Our method demonstrates robustness to color variations and accommodates local patch editing, ensuring stable 3D facial applications even from less meticulous makeup edits. Yang-Res requires extra care in editing due to its over-fitting of modified images.

\vskip0.5\baselineskip
\noindent \textbf{3D makeup transfer.}
The estimated makeup can be utilized for a 3D makeup transfer application. As shown in Fig.~\ref{fig:result_transfer}, Yang-Ext~\cite{MakeupExtract} struggles with faces in large poses during estimation, leading to partial makeup loss in the makeup transfer application. The Yang-Res~\cite{MakeupExtract}, being tightly coupled with the original makeup images, fails to facilitate makeup transfer. Conversely, our methods, benefiting from superior 3D makeup estimation outcomes, render the application of 3D makeup transfer both precise and effective.

\vskip0.5\baselineskip
\noindent \textbf{3D makeup interpolation.}
To better visualize, we transfer the interpolated makeup to other 3D faces. Figs.~\ref{fig:teaser} and \ref{fig:result_interpolate} illustrate our experiments with makeup interpolation using the makeup prior model. Beyond the alpha matte adjustment proposed by the Yang-Ext~\cite{MakeupExtract}, our approach leverages the inherent characteristics of the prior model, performing linear interpolation on the coefficients. Taking the StyleGAN2 prior model as an example, we demonstrate two outcomes: the top row in Fig.~\ref{fig:result_interpolate} shows the results of a full makeup interpolation between the estimated makeup coefficients of two reference images, with the noticeable blending of eyeshadow and lipstick. The bottom row illustrates our capability for partial makeup interpolation through style mixing~\cite{pSp}. This is achieved by retaining the selected layers of the coefficient from reference A and interpolating the remaining layers from reference B. The lipstick color is preserved while only the eyeshadow changes. We hypothesize that this is attributable to the StyleGAN2 makeup prior model's training, which allows for more effective disentanglement of the coefficient~\cite{StyleGAN}. This represents a significant advantage over the PCA model. We intend to further investigate this feature, with the potential to apply it to applications such as makeup manipulation.

\subsection{Computational Time} 
\label{sec:computation_time}
We measured the computational time for 3D makeup estimation. The Yang-Res and Yang-Ext methods~\cite{MakeupExtract} require 56.12s and 63.13s, respectively. On the other hand, our StyleGAN2-based method demonstrates a substantial speed improvement, with the inference time at 18.13s—about 3 times faster—where the makeup estimation network contributes 1.66s and the refinement process 16.47s. Our PCA-based approach achieves a remarkable 180-fold speed increase, clocking just 0.31s. We believe that while the StyleGAN2-based method offers superior quality in makeup estimation, the high-speed PCA method remains valuable for real-time situations.

\subsection{Ablation Studies}
As shown in Fig.~\ref{fig:ablation}, we conducted ablation studies on the makeup prior models (PCA and StyleGAN) to compare the outcomes with and without the use of the makeup consistency module, as well as the additional refinement process. The top row indicates that the utilization of $\mathcal{L}_{cyc}$ leads to enhanced robustness. The bottom row shows that while refinement tends to introduce artifacts in the PCA model, it enhances the precision of the StyleGAN2 model. As a trade-off, even with the implementation of loss functions $\mathcal{L}_{sym}$ and $\mathcal{L}'_{reg}$ in our refinement process, the absence of the makeup consistency module may still increase the coupling between makeup and facial features. 
Fig.~\ref{fig:ablation_refine} shows the experiments of the loss functions $\mathcal{L}_{sym}$ and $\mathcal{L}'_{reg}$. The absence of these loss functions leads to over-fitting of the input image, resulting in artifacts. 
Further details of ablation studies will be presented in the supplementary materials.

\section{Conclusions}
We have presented two makeup prior models: a PCA-based linear model for time efficiency and a StyleGAN2-based model for high-fidelity detail representation in 3D facial makeup applications. Our method addresses the challenges of \revised{self-occluded faces}, enhancing robustness and reducing computation time. Demonstrated across multiple applications, our methods show improved accuracy and versatility in 3D facial makeup estimation, reconstruction, user-friendly makeup editing, transfer, and interpolation. Future work will explore the decoupling capabilities of the StyleGAN2 model for makeup manipulation.    
{
    \small
    \bibliographystyle{ieeenat_fullname}
    \bibliography{main}
}

\clearpage
\setcounter{page}{1}
\maketitlesupplementary

\renewcommand{\thefigure}{S.\arabic{figure}}
\renewcommand{\thetable}{S.\arabic{table}}
\renewcommand{\theequation}{S.\arabic{equation}}

\appendix
\revised{In this supplemental material, 
we first briefly review the two methods proposed by Yang \etal~\cite{MakeupExtract} in Sec.~\ref{sec:ext_and_pca}; the texture decomposition-based 3D facial makeup extraction and the optimization-based residual makeup model fitting, which we named ``Yang-Ext'' and ``Yang-Res'', respectively. Then, we compare the Yang-Res and our PCA model in Sec.~\ref{sec:difference_pca}, which aims for a better understanding of the differences. }

For the supplemental experiments, we first provide additional details on the ablation studies in Sec.~\ref{sec:ablation}, as promised in the main text. This includes two comparisons; one is an experiment on the refinement process using inferred coefficients for initialization in Sec.~\ref{sec:initialization}, and the other is the details of quantitative experiments in Sec.~\ref{sec:quantitative}.
Subsequently, we present more results related to makeup extraction, 3D face reconstruction, makeup interpolation, and makeup transfer in Sec.~\ref{sec:additional_results}.

\section{\revised{Previous Makeup Estimation Methods}}
\label{sec:1}
\subsection{Yang-Ext and Yang-Res}
\label{sec:ext_and_pca}
Yang \etal.~\cite{MakeupExtract} tackled the problem of 3D-aware makeup extraction for the first time, in which they proposed two variants, i.e., what we call ``Yang-Ext'' and ``Yang-Res''.
We first explain the texture decomposition-based method, ``Yang-Ext'', which consists of the following four coarse-to-fine steps: 

\begin{description}
    \item Step 1 servers as coarse facial texture extraction via regression-based inverse rendering using a 3D morphable model (3DMM). The facial textures are represented as UV maps for the 3DMM. This step extends the method by Deng \etal~\cite{deng2020accurate} to estimate not only face shape, albedo, and diffuse shading but also specular shading. The rendering method for this 3DMM reconstruction includes both the Lambertian and Blinn-Phong reflection models. The pre-trained 3D face reconstruction network ${E}_\text{FLAME}$ in our method leverages the same network as theirs.
    \item Step 2 employs inpainting in UV maps. Their method tries to extract detailed information by fitting the 3DMM to facial images, but the images might be partially hidden due to self-occlusion. It thus utilize a face UV inpainting technique~\cite{dsd-gan}.
    \item Step 3 refines the inpainted coarse textures via optimization to obtain a high-resolution albedo map that contains bare skin and makeup without the effect of illumination.
    \item Step 4 separates the high-resolution albedo map into bare skin, makeup, and alpha matte based on makeup transfer. The combination of these decomposed textures follows Eq.~(\ref{eq:pca_M}) and Eq.~(\ref{eq:blending})
\end{description}

As an application, they perform principal component analysis on the extracted makeup textures to create a three-channel morphable makeup model follows Eq.~(\ref{eq:addition}). Therefore, they can implement an optimization-based fitting of the makeup model to makeup images, which we named ``Yang-Res''.

The entire method of Yang-ext is a long pipeline, making the algorithm unstable and time-consuming.
The refinement of albedo in Step 3 is performed using the coarse albedo obtained in Step 1, which does not contain makeup, affecting subsequent accuracy.

\subsection{Differences between Yang-Res and Our PCA}
\label{sec:difference_pca}

Briefly, Yang-Res has the following key limitations:
\begin{enumerate}
\item \textbf{Undecoupled approach}: It models makeup as a residual via subtraction (see Sec.~7.1 in their paper~\cite{MakeupExtract}). The residual closely ties the makeup to its corresponding bare skin. 
This approach is neither easily generalizable nor versatile for transferring to different faces, as it may carry excessive residuals that do not match the new face's features, as shown in Fig.~\ref{fig:result_transfer}.
\item \textbf{Analysis-by-synthesis optimization}: The optimization incurs significant computational costs due to several hundreds of iterations, as opposed to the single forward pass of our regression network.
\item \textbf{Lacks makeup-specific regularization}: As shown in Fig.~\ref{fig:result_edit}, without the makeup-specific regularization to constrain PCA parameters, makeup becomes unconstrained and thus overfits the input images.
\end{enumerate}
Our PCA addresses these issues through the following improvements:
\begin{enumerate}
\item \textbf{Alpha blending model}: 
We adopt the same makeup model as Yang-Ext, \ie, the alpha blending model, which
reduces the risk of overfitting by independently estimating makeup bases and opacity, thereby decreasing reliance on specific features.
Our $\mathbf{M}_b$ and $\mathbf{M}_a$ correspond to Yang-Ext's $\mathbf{D}_{m}^{m}$ and $(1 - \mathbf{A})$, respectively.
\item \textbf{Regression network}: We trained a regression network to accelerate the inference.
\item \textbf{Makeup consistency module}: We designed a novel architecture to regularize makeup PCA parameters. This module transfers makeup across different identities, expressions, and poses while maintaining cycle consistency. This enhances the disentanglement of makeup and bare skin, enabling us to handle various scenarios, such as occluded faces resulting from large poses.
\end{enumerate}

\noindent It is notable that even though we could employ Yang-Res to implement the acceleration in inference (Item 2), Yang-Res still suffers from the shortcomings of Items 1 and 3.

\section{Additional Details on Ablation Studies}
\label{sec:ablation}

\subsection{Coefficient Initialization for Refinement}
\label{sec:initialization}
\begin{figure}[ht]
    \centering
    \includegraphics[width=0.99\linewidth]{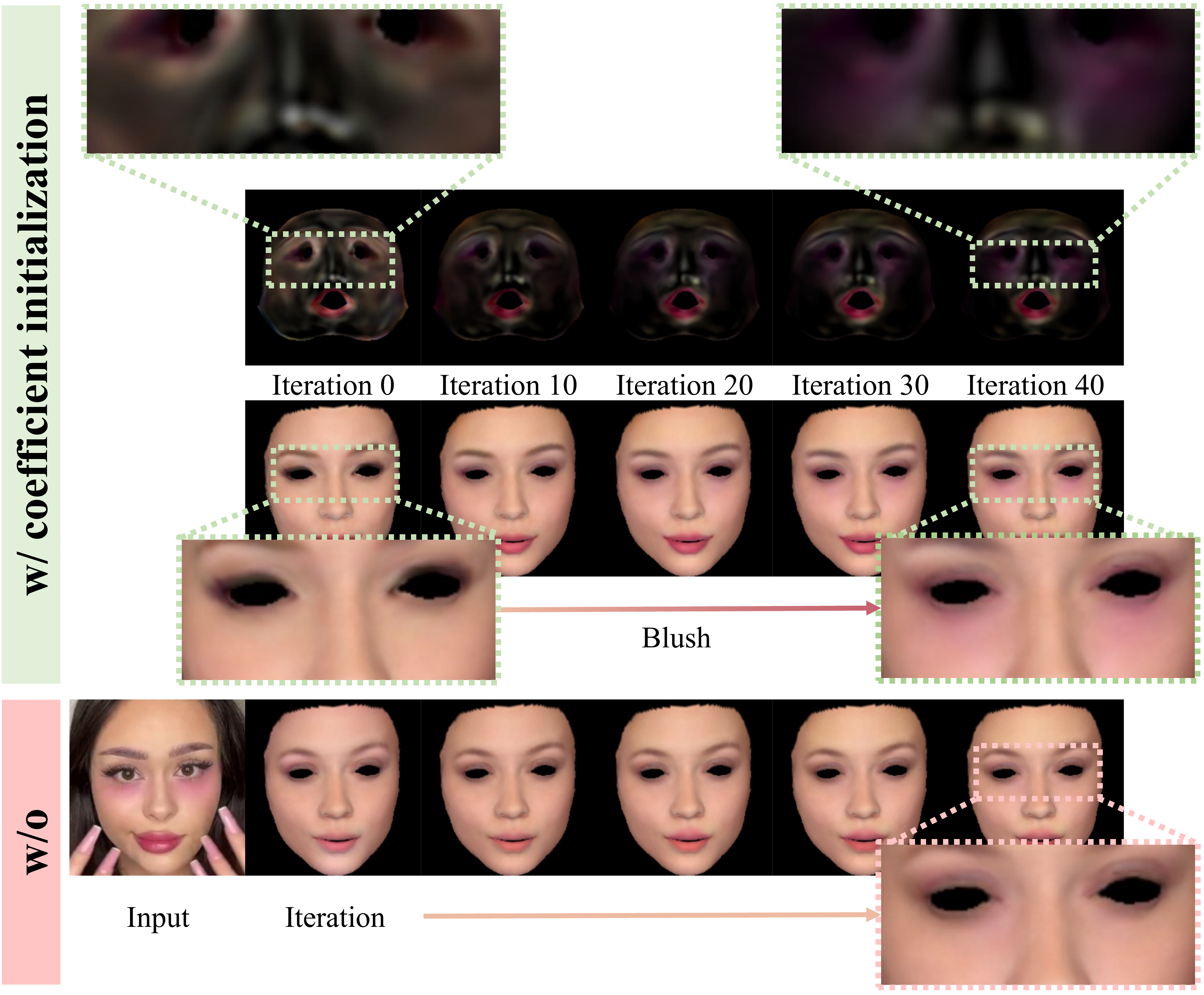}
    \caption{Qualitative comparison of
    makeup refinement with and without initialization of coefficients inferred by the makeup estimation network.
    With initialization, detailed makeup features can be achieved with minimal iterations of refinement, as exemplified by the enhancement of blush details.}
    \label{fig:supplement_refinement}
\end{figure}

In Fig.~\ref{fig:supplement_refinement}, 
we demonstrate that initialization with inferred coefficients obtained from the makeup estimation network has a large impact on makeup refinement.
For this demonstration, we chose blush as an example. It is important to note that without using these initial values, the quality of makeup estimation does not show significant improvement.

\subsection{Details of Quantitative Evaluations}
\label{sec:quantitative}
\begin{figure}[t]
    \centering
    \includegraphics[width=0.99\linewidth]{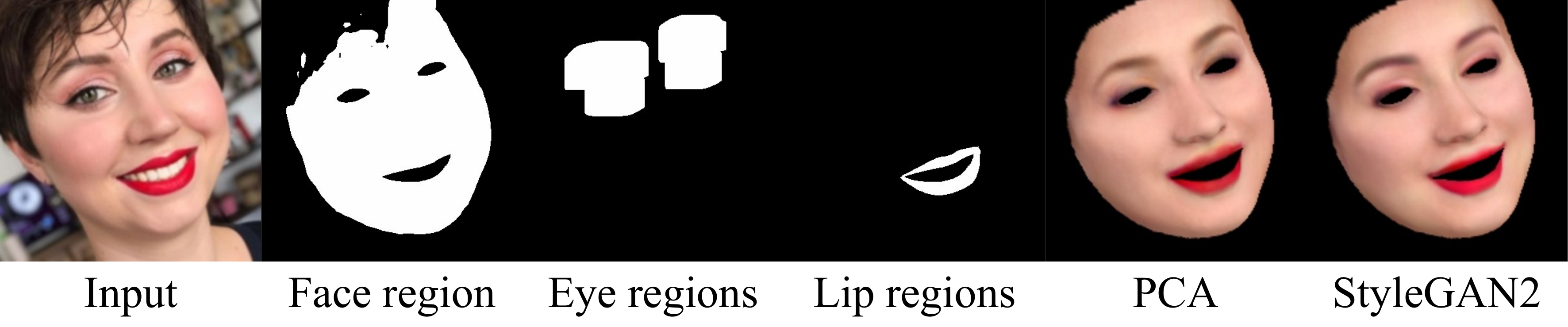}
    \caption{Examples of segmented masks for quantitative experiments.}
    \label{fig:supplement_region}
\end{figure}

In the quantitative evaluations, we compare the differences between input images and the rendering results of 3D face reconstruction. 
Since there is no ground truth data to evaluate the makeup of invisible face regions, our experiments are based on visible areas.
As shown in Fig.~\ref{fig:supplement_region}, we first perform segmentation on the input images to obtain masks for the face, eyes, and lips. We employ morphological dilation with a $15\times15$ kernel, iterating three times, to expand the regions of the eyes and eyebrows. This approach ensures coverage of most of the eyeshadow area. Regarding the metrics, 
we use the eye regions and lip regions for the Histogram Matching (HM) metric, while we use the face region for the other metrics.

\begin{figure}[t]
    \centering
    \includegraphics[width=0.99\linewidth]{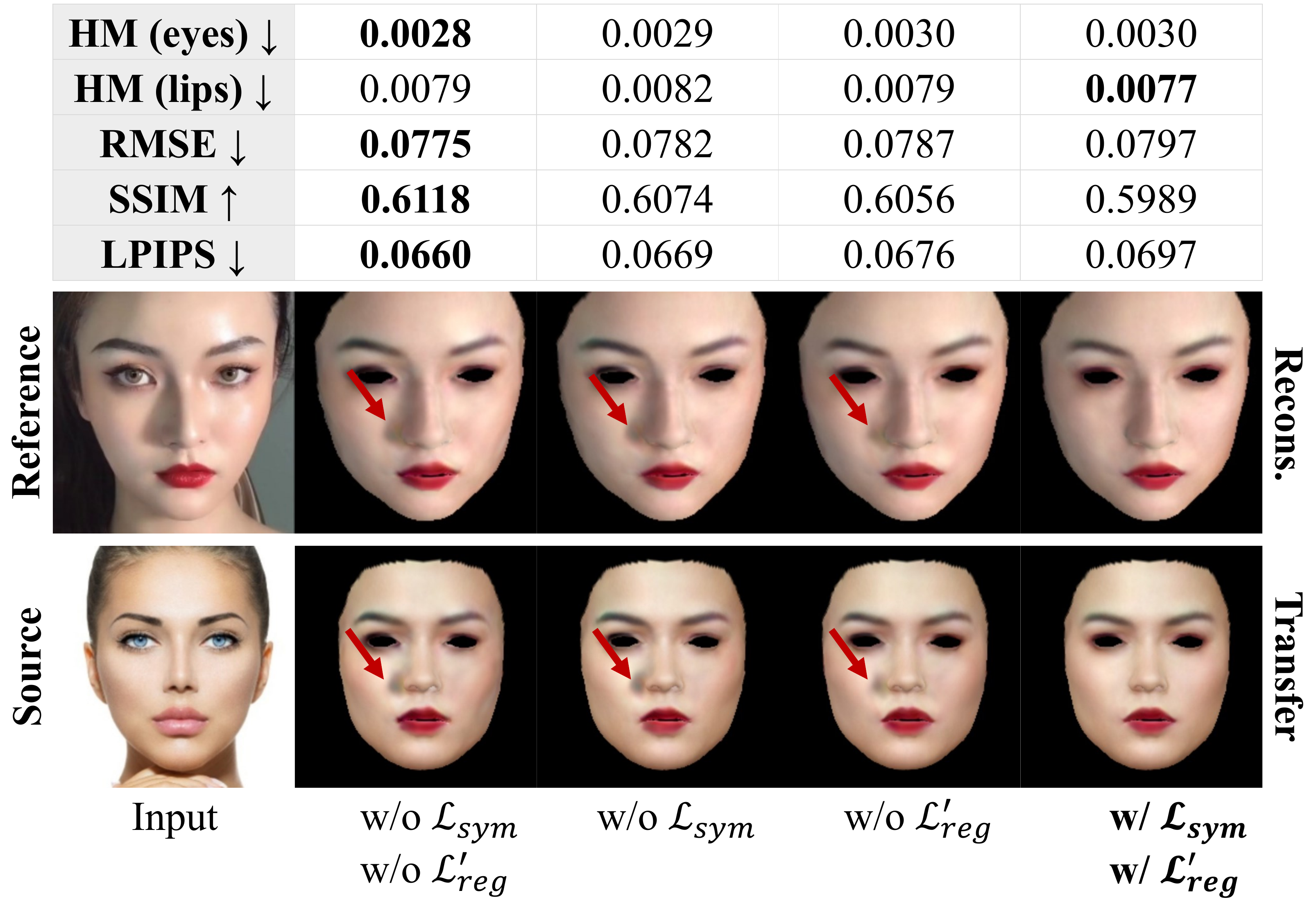}
    \caption{Example of quantitative and qualitative comparison for loss functions. The improved quantitative metrics do not necessarily benefit downstream makeup-related tasks.}
    \label{fig:supplement_refine_transfer}
\end{figure}

Tab.~\ref{tab:suppl} presents a quantitative ablation study on the loss functions. While $\mathcal{L}_{cyc}$, $\mathcal{L}_{sym}$, and $\mathcal{L}'_{reg}$ induce some fluctuations in quantitative metrics, they actually help the model avoid overfitting, qualitatively resulting in more natural and realistic visual makeup effects. In other words, the impact of these loss functions may not be easily captured by traditional quantitative measures but is significant in terms of visual outcomes. Therefore, we consider qualitative and quantitative evaluations as complementary assessment methods. For instance, as shown in Fig.~\ref{fig:supplement_refine_transfer}, overfitting to the shadows beside the nose as part of the makeup patterns might enhance quantitative metrics. However, this can hinder the effectiveness of makeup transfer across different individuals. The same rationale applies to user-friendly editing scenarios.

\begin{table*}[!t]
\caption{\textbf{Comparative analysis of loss functions in 3DMM-based 3D face reconstruction.} Values in \textbf{bold} represent the best results.}
\begin{center}
\resizebox{0.95\textwidth}{!}{
{\small
\begin{tabular}{lcccccccccc}
\toprule
\multicolumn{1}{l}{ } & \multicolumn{5}{c}{\textbf{Wild}~\cite{PSGAN}} & \multicolumn{5}{c}{\textbf{BeautyFace}~\cite{BeautyREC}} \\
\cmidrule(lr){2-6} \cmidrule(lr){7-11}
Method & HM(eyes)$\downarrow$ & HM(lips)$\downarrow$ & RMSE$\downarrow$ & SSIM$\uparrow$ & LPIPS$\downarrow$ & HM(eyes)$\downarrow$ & HM(lips)$\downarrow$ & RMSE$\downarrow$ & SSIM$\uparrow$ & LPIPS$\downarrow$\\ 
\midrule
w/o $\mathcal{L}_{cyc}$ (PCA) & 0.0041 & 0.0078 & \textbf{0.0604} & \textbf{0.6119} & \textbf{0.0667} & 0.0035 & \textbf{0.0076} & \textbf{0.0684} & \textbf{0.5021} & \textbf{0.0718}\\
\textbf{w/ $\mathcal{L}_{cyc}$ (PCA)} & 0.0041 & 0.0078 & 0.0609 & 0.6111 & 0.0681 & 0.0035 & 0.0078 & 0.0690 & 0.5013 & 0.0733\\
\midrule
w/o $\mathcal{L}_{cyc}$ (StyleGAN2 w/o Refine) & 0.0042 & 0.0086 & 0.0634 & 0.6078 & 0.0707 & 0.0036 & 0.0087 & 0.0707 & 0.5021 & 0.0763\\
\textbf{w/ $\mathcal{L}_{cyc}$ (StyleGAN2 w/o Refine)} & 0.0042 & \textbf{0.0083} & \textbf{0.0618} & \textbf{0.6091} & \textbf{0.0685} & \textbf{0.0035} & \textbf{0.0084} & \textbf{0.0695} & \textbf{0.5031} & \textbf{0.0741}\\
\midrule
w/o $\mathcal{L}_{sym}$ w/o $\mathcal{L}'_{reg}$ (StyleGAN2) & \textbf{0.0034} & \textbf{0.0071} & \textbf{0.0464} & \textbf{0.6386} & \textbf{0.0534} & \textbf{0.0029} & \textbf{0.0066} & \textbf{0.0621} & \textbf{0.5293} & \textbf{0.0601}\\
w/o $\mathcal{L}_{sym}$ (StyleGAN2) & 0.0035 & 0.0072 & 0.0488 & 0.6313 & 0.0559 & 0.0030 & 0.0067 & 0.0633 & 0.5206 & 0.0628\\
w/o $\mathcal{L}'_{reg}$ (StyleGAN2) & 0.0035 & 0.0072 & 0.0493 & 0.6309 & 0.0596 & 0.0030 & 0.0067 & 0.0637 & 0.5214 & 0.0654\\
\textbf{\textbf{w/ $\mathcal{L}_{sym}$ w/ $\mathcal{L}'_{reg}$} (StyleGAN2)} & 0.0036 & 0.0073 & 0.0517 & 0.6240 & 0.0608 & 0.0031 & 0.0068 & 0.0650 & 0.5134 & 0.0673\\
\bottomrule
\end{tabular}
}}
\end{center}
\label{tab:suppl}
\end{table*}

\section{Additional Results}
\label{sec:additional_results}
\begin{figure*}[t]
    \centering
    \includegraphics[width=0.95\linewidth]{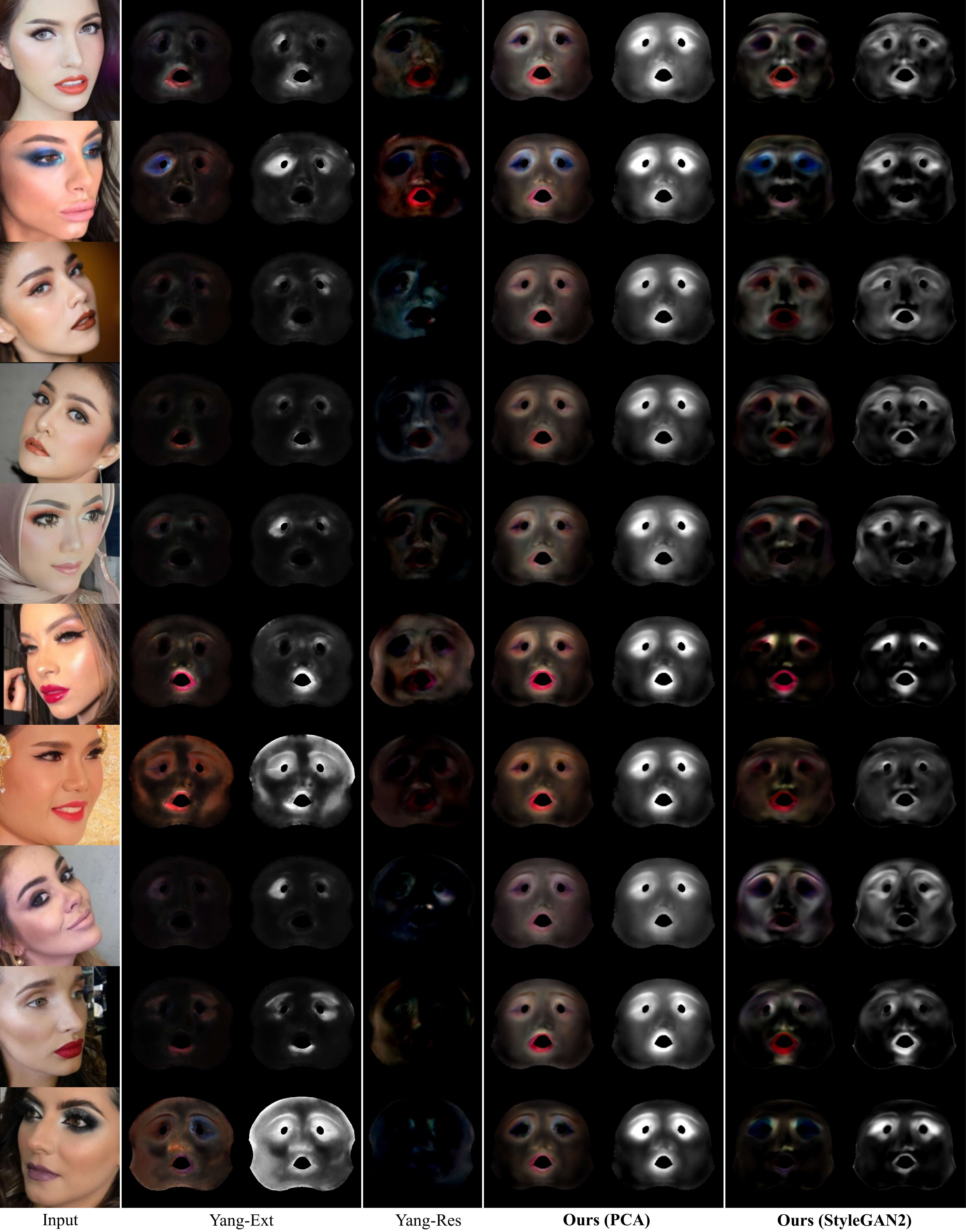}
    \caption{\textbf{Comparison to previous work.} Our methods (PCA and StyleGAN2) outperform both Yang-Ext and Yang-Res~\cite{MakeupExtract}, which show limitations in handling self-occluded faces.}
    \label{fig:supplement_estimation_1}
\end{figure*}

\begin{figure*}[t]
    \centering
    \includegraphics[width=0.95\linewidth]{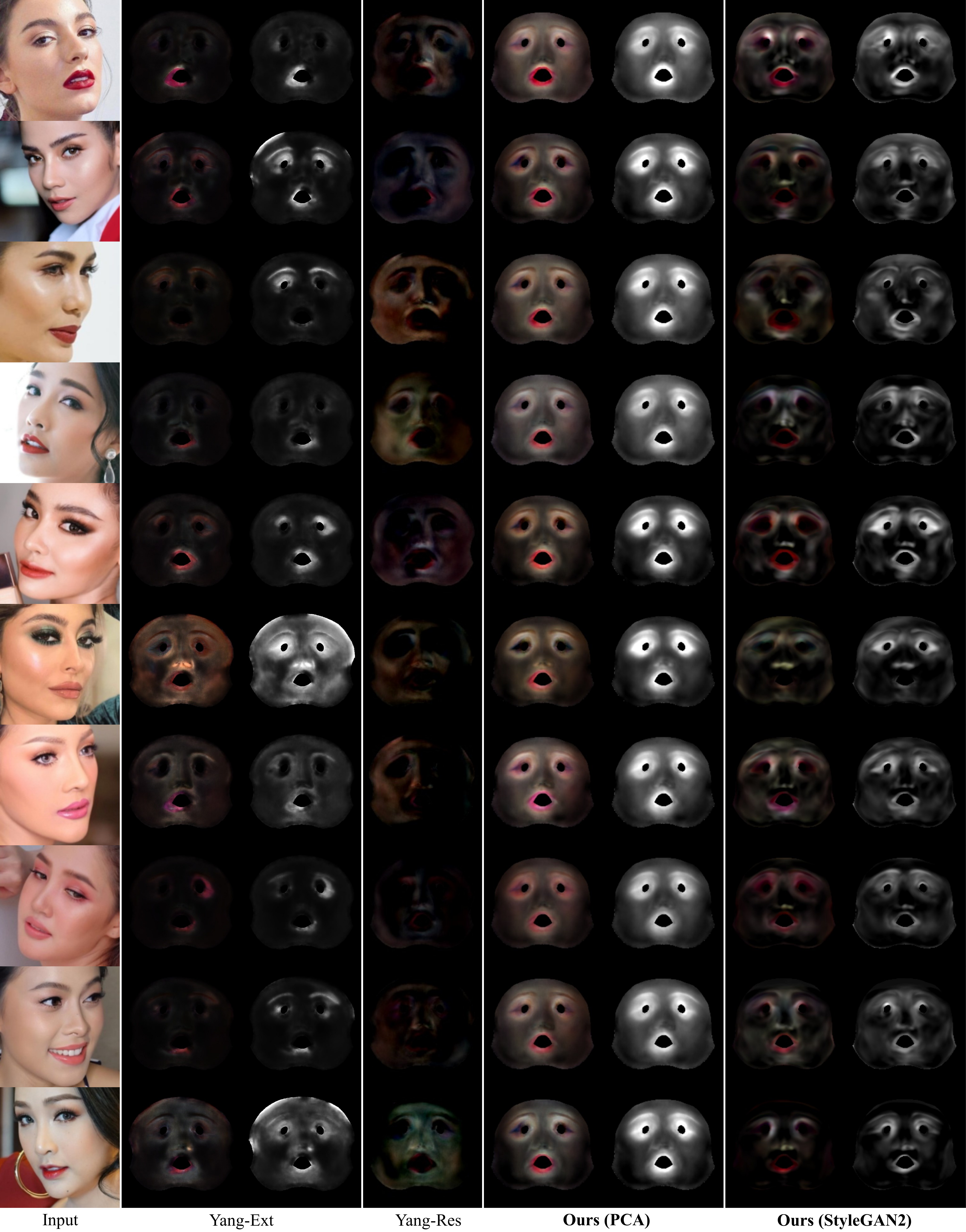}
    \caption{\textbf{Additional comparison to previous work.} Our methods (PCA and StyleGAN2) outperform both Yang-Ext and Yang-Res~\cite{MakeupExtract}, which show limitations in handling self-occluded faces.}
    \label{fig:supplement_estimation_2}
\end{figure*}

\begin{figure*}[t]
    \centering
    \includegraphics[width=0.98\linewidth]{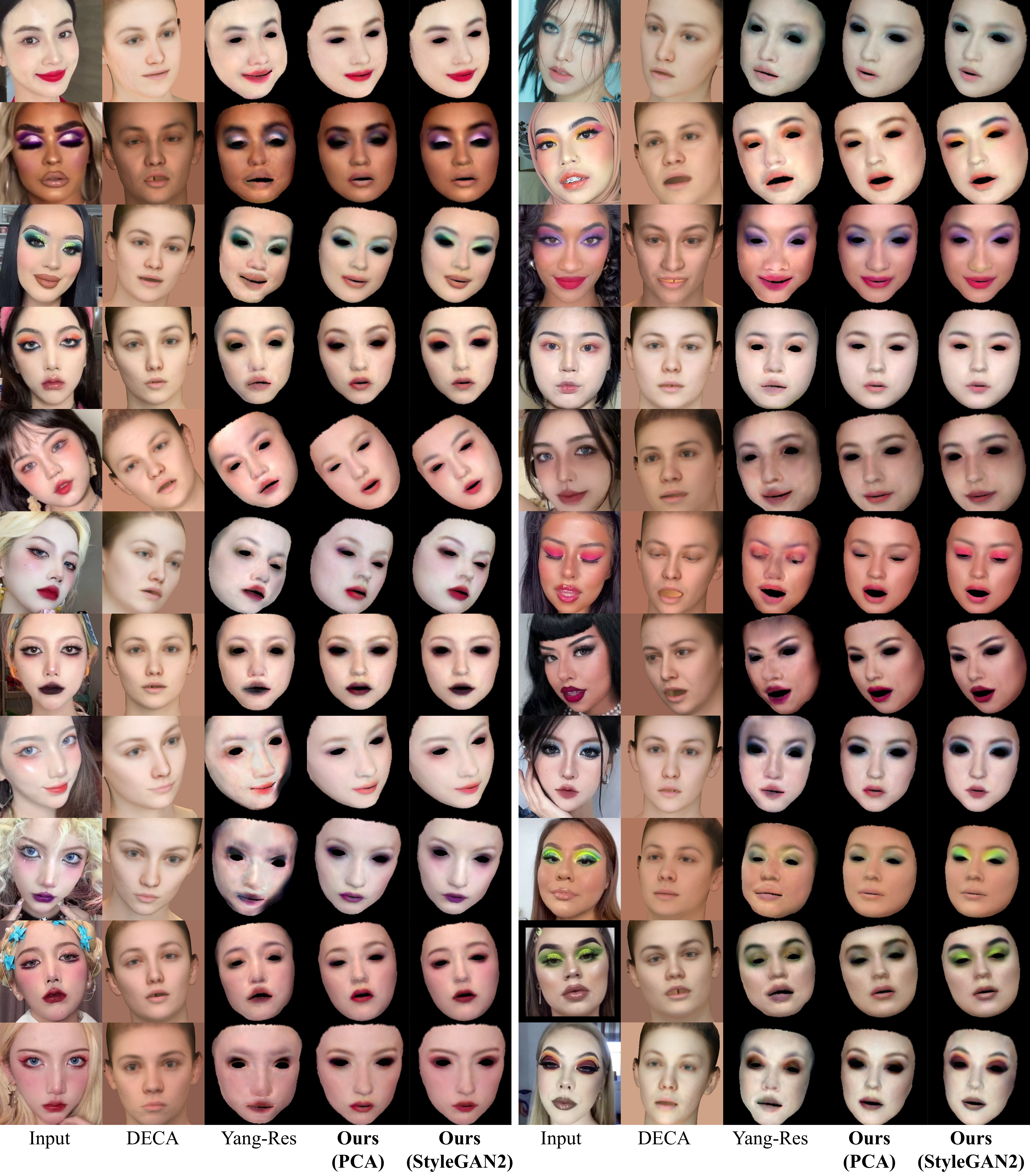}
    \caption{\textbf{Comparison with 3D face reconstruction methods using our makeup prior models.} Our methods successfully reconstruct facial makeup. Specifically, our PCA model is capable of broadly recovering makeup colors, while our StyleGAN2 model achieves precise replication of complex makeup features, such as blush and gradational eyeshadow.}
    \label{fig:supplement_recons_1}
\end{figure*}

\begin{figure*}[t]
    \centering
    \includegraphics[width=0.98\linewidth]{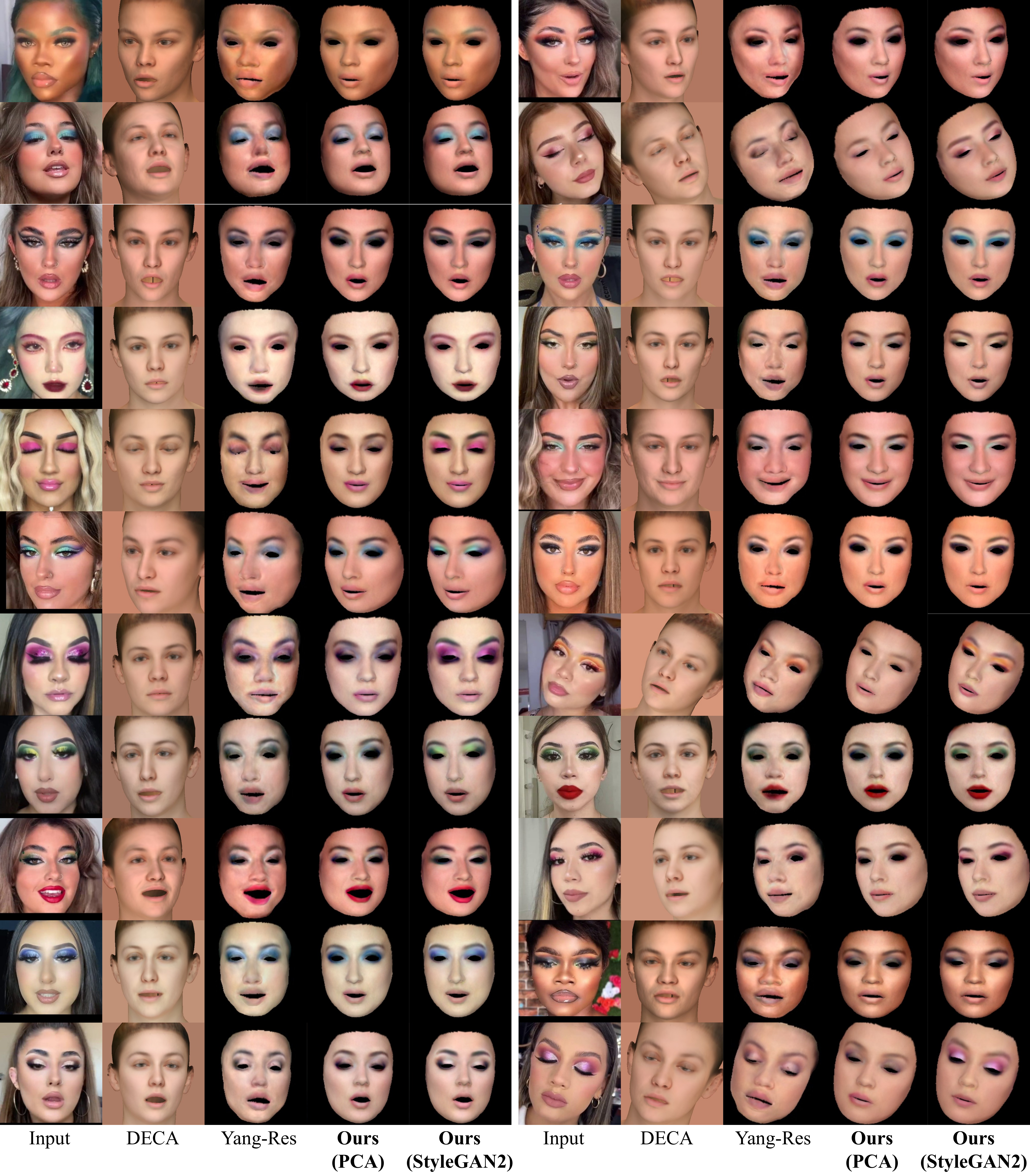}
    \caption{\textbf{Additional comparison with 3D face reconstruction methods using our makeup prior models.} Our methods successfully reconstruct facial makeup. Specifically, our PCA model is capable of broadly recovering makeup colors, while our StyleGAN2 model achieves precise replication of complex makeup features, such as blush and gradational eyeshadow.}
    \label{fig:supplement_recons_2}
\end{figure*}

\begin{figure*}[t]
    \centering
    \includegraphics[width=0.99\linewidth]{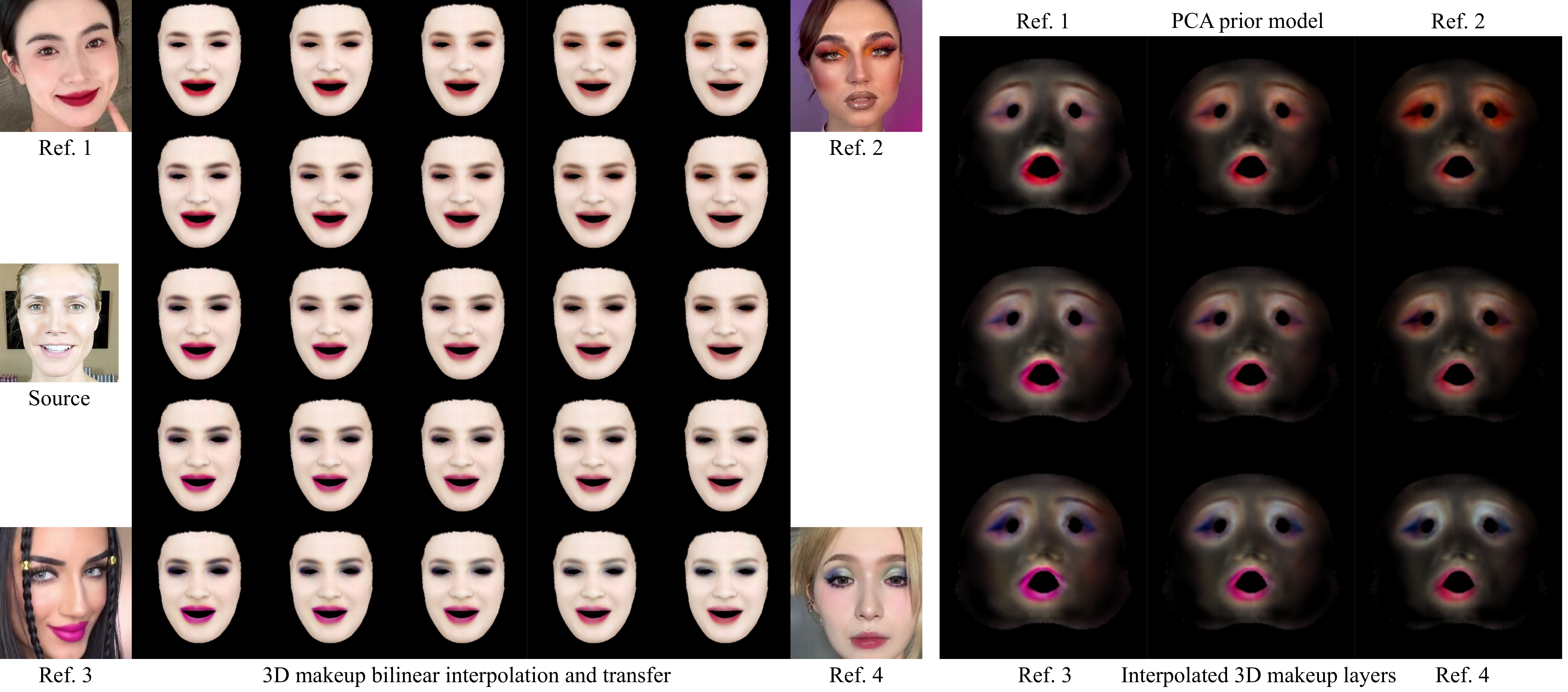}
    \caption{\textbf{Results of 3D makeup interpolation and transfer using our PCA model.} Left: the bilinear makeup interpolation between four makeup styles. Right: the estimated makeup layers and the interpolated results using the makeup coefficient $\mathbf{\upsilon}$.}
    \label{fig:supplement_interpolate_pca}
\end{figure*}

\begin{figure*}[t]
    \centering
    \includegraphics[width=0.99\linewidth]{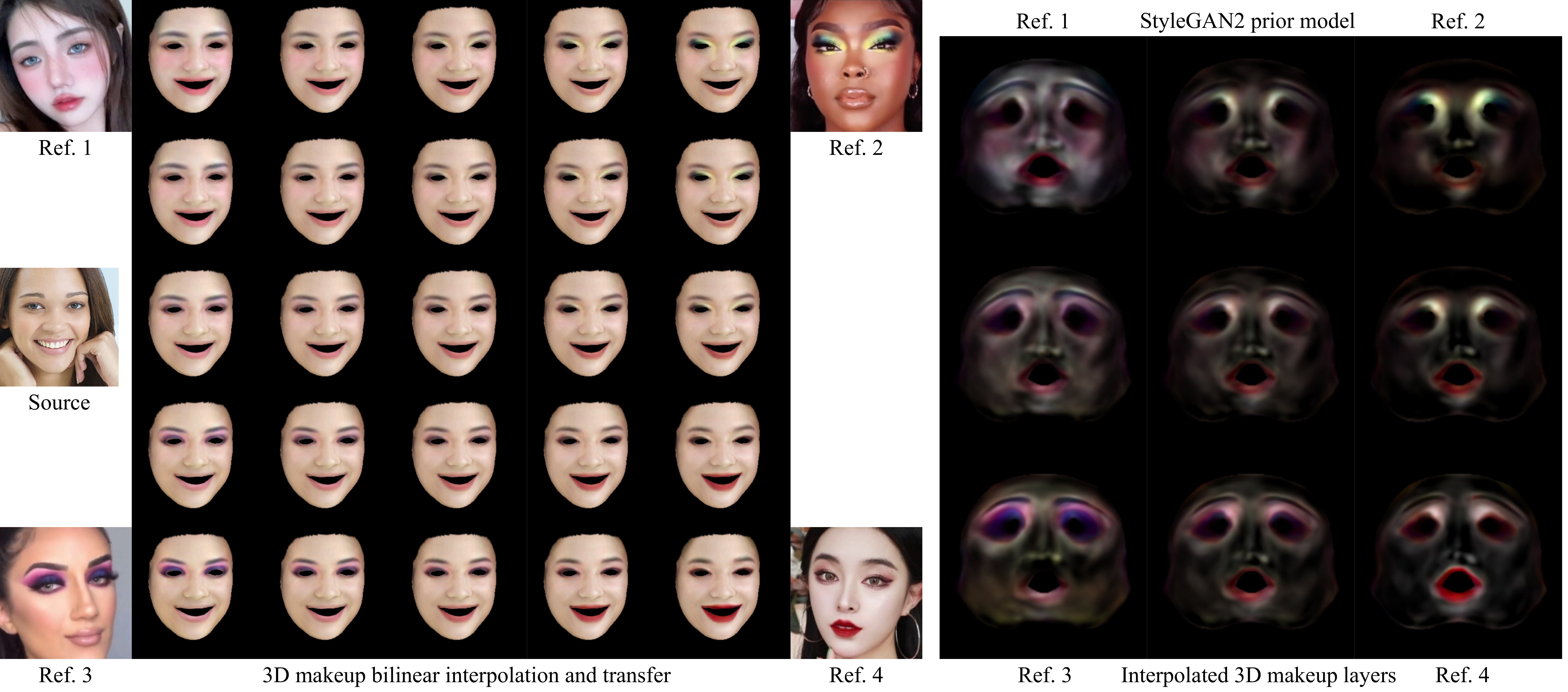}
    \caption{\textbf{Results of 3D makeup interpolation and transfer using our StyleGAN2 model.} Left: the bilinear makeup interpolation between four makeup styles. Right: the estimated makeup layers and the interpolated results using the makeup coefficient $\textbf{w}$.}
    \label{fig:supplement_interpolate_stylegan}
\end{figure*}

\revised{To the best of our knowledge, apart from Yang \etal, no other relevant methods address makeup for 3DMM or model-based single image 3D makeup estimation. Hence, we primarily compared against Yang \etal as a baseline method.} Figs.~\ref{fig:supplement_estimation_1} and \ref{fig:supplement_estimation_2} show additional comparative results of makeup estimation, particularly in scenarios involving \revised{self-occluded faces}. 
Figs.~\ref{fig:supplement_recons_1} and \ref{fig:supplement_recons_2} display additional comparative results for model-based 3D facial reconstruction with makeup. We selected challenging makeup styles, such as uncommon colors and gradational eyeshadows, blushes, and lipsticks for comparison. 
We use DECA~\cite{DECA} as a reference that lacks a makeup prior model.
As shown in Figs.~\ref{fig:supplement_interpolate_pca} and \ref{fig:supplement_interpolate_stylegan}, we demonstrate the makeup interpolation and transfer using makeup coefficients for both the PCA and StyleGAN2 models. It can be observed that we can achieve multi-interpolation by blending makeup coefficients. In our examples, we utilize bilinear interpolation for illustration.

\end{document}